\documentclass{article}

\usepackage{microtype}
\usepackage{graphicx}
\usepackage{booktabs} 
\usepackage[dvipsnames]{xcolor}

\usepackage{xspace}
\usepackage[shortlabels]{enumitem}
\usepackage{subcaption}

\usepackage{hyperref}

\newcommand{\ouralgo}{RLPD\xspace}

\usepackage[accepted]{icml2023}

\usepackage{amsmath}
\usepackage{amssymb}
\usepackage{mathtools}
\usepackage{amsthm}
\usepackage{wrapfig}
\usepackage{algorithm}
\usepackage{amsfonts}

\usepackage[capitalize,noabbrev]{cleveref}

\theoremstyle{plain}

\theoremstyle{definition}

\theoremstyle{remark}

\makeatletter
\def\mathcolor#1#{\@mathcolor{#1}}
\def\@mathcolor#1#2#3{%
  \protect\leavevmode
  \begingroup
    \color#1{#2}#3%
  \endgroup
}
\makeatother

\usepackage[textsize=tiny]{todonotes}

\icmltitlerunning{Efficient Online RL with Offline Data}

\begin{document}

\twocolumn[

\icmltitle{Efficient Online Reinforcement Learning with Offline Data}



\icmlsetsymbol{equal}{*}

\begin{icmlauthorlist}
\icmlauthor{Philip J. Ball}{equal,ox}
\icmlauthor{Laura Smith}{equal,ber}
\icmlauthor{Ilya Kostrikov}{equal,ber}
\icmlauthor{Sergey Levine}{ber}
\end{icmlauthorlist}

\icmlaffiliation{ox}{University of Oxford}
\icmlaffiliation{ber}{UC Berkeley}

\icmlcorrespondingauthor{Philip J. Ball}{\href{mailto:ball@robots.ox.ac.uk}{ball@robots.ox.ac.uk}}
\icmlcorrespondingauthor{Laura Smith}{\href{mailto:smithlaura@berkeley.edu}{smithlaura@berkeley.edu}}
\icmlcorrespondingauthor{Ilya Kostrikov}{\href{mailto:kostrikov@berkeley.edu}{kostrikov@berkeley.edu}}

\icmlkeywords{Machine Learning, ICML}

\vskip 0.3in
]
\printAffiliationsAndNotice{\icmlEqualContribution} 

\begin{abstract}
Sample efficiency and exploration remain major challenges in online reinforcement learning (RL). A powerful approach that can be applied to address these issues is the inclusion of offline data, such as prior trajectories from a human expert or a sub-optimal exploration policy. Previous methods have relied on extensive modifications and additional complexity to ensure the effective use of this data. Instead, we ask: \emph{can we simply apply existing off-policy methods to leverage offline data when learning online?}  
In this work, we demonstrate that the answer is yes; however, a set of minimal but important changes to existing off-policy RL algorithms are required to achieve reliable performance.
We extensively ablate these design choices, demonstrating the key factors that most affect performance, and arrive at a set of recommendations that practitioners can readily apply, whether their data comprise a small number of expert demonstrations or large volumes of sub-optimal trajectories. We see that correct application of these simple recommendations can provide a $\mathbf{2.5\times}$ improvement over existing approaches across a diverse set of competitive benchmarks, with no additional computational overhead. We have released our code here: \href{https://github.com/ikostrikov/rlpd}{\texttt{github.com/ikostrikov/rlpd}}.
\end{abstract}

\section{Introduction}
Deep reinforcement learning (RL) has achieved success in a number of complex domains, such as Atari \citep{mnih2015humanlevel} and Go \citep{alphago}, as well as real-world applications like Chip Design \citep{Mirhoseini2021AGP} and Human Preference Alignment \citep{ouyang2022training}. In many of these settings, strong RL performance is predicated on having large amounts of online interaction with an environment, which is usually made feasible through the use of simulators.
In real-world problems, however, we are often confronted with scenarios where samples are expensive, and furthermore, rewards are sparse, often exacerbated by high dimensional state and action spaces.

\begin{figure}[t]
    \centering
    \vspace{-3mm}
    \includegraphics[width=0.4\textwidth]{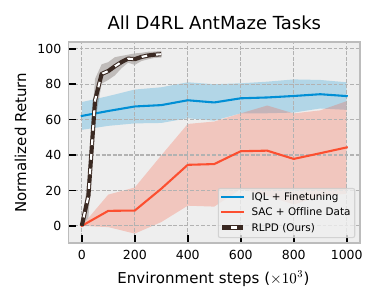} \\
    \vspace{-6mm}
    \caption{\small{Our approach, \ouralgo, extends standard off-policy RL and achieves reliable state-of-the-art online performance on a number of tasks using offline data. Here we show the difficult D4RL AntMaze domain (10 seeds, 1 std. shaded), averaged over all 6 tasks. We run \ouralgo for 300k steps due to early convergence.}}
    \label{fig:ourmethod}
    \vspace{-5mm}
\end{figure}

One promising way to resolve this issue is via the inclusion of data generated by a previous policy or human expert when training deep RL algorithms (often referred to as \emph{offline data} \citep{offlineRL}), as evidenced theoretically \citep{wagenmaker2022leveraging,song2023hybrid} and in real-world examples by \citet{Cabi2019ScalingDR,awac,awopt2021corl}. This can alleviate challenges due to sample efficiency and exploration by providing the algorithm with an initial dataset to ``kick-start" the learning process, either in the form of high-quality expert demonstrations, or even low-quality but high-coverage exploratory trajectories. This also provides us with an avenue to leverage large pre-collected datasets in order to learn useful policies.

Some prior work has focused on using this data through pre-training, while other approaches introduce constraints when training online to handle issues with distribution shift. However, each approach has its drawbacks, such as additional training time and hyperparamters, or limited improvement beyond the behavior policy respectively. Taking a step back, we note that \emph{standard off-policy algorithms} should be able to take advantage of this offline data, and furthermore issues with distribution shift should be alleviated in this setting, as we can explore the environment online. Thus far however, such methods have seen limited success in this problem setting. Therefore, in this work, we ask the following question: \emph{can we simply apply existing off-policy methods to leverage offline data when learning online, without offline RL pre-training or explicit imitation terms that privilege the prior offline data?}

Through a set of thorough experiments on a collection of widely studied benchmarks, we show that the answer to this question is yes. However, na\"ively applying existing online off-policy RL algorithms can result in comparatively poor performance, as we see in \cref{fig:ourmethod} comparing `SAC + Offline Data' with `IQL + Finetuning'. Instead, a minimal set of key \emph{design choices} must be taken into consideration to ensure their success. Concretely, we first introduce a remarkably simple approach to sampling the offline data, which we call ``symmetric sampling", that performs well over a large variety of domains with no hyperparameter tuning. Then, we see that in complex settings (e.g., sparse reward, low volume of offline data, high dimensionality, etc.), it is vital that value functions are prevented from over-extrapolation. To this end, we provide a novel perspective on how Layer Normalization \cite{layernorm} implicitly prevents catastrophic value over-extrapolation, thereby greatly improving sample-efficiency and stability in many scenarios, while being a minimal modification to existing approaches. Then, to improve the rate at which the offline data are utilized, we incorporate and compare the latest advances in sample-efficient model-free RL, and find that large ensembles are remarkably effective across a variety of domains. Finally, we identify and provide evidence that key design choices in recent RL literature are in fact environment sensitive, showing the surprising result that environments which share similar properties in fact require entirely different choices, and recommend a \emph{workflow for practitioners} to accelerate their application of our insights to new domains.

We demonstrate that our final approach, which we call \textbf{\ouralgo} (\textbf{R}einforcement \textbf{L}earning with \textbf{P}rior \textbf{D}ata) outperforms previously reported results, as we see in \cref{fig:ourmethod}, on many competitive domains, sometimes by $\mathbf{2.5\times}$. Crucially, as our changes are minimal, we maintain the attractive properties of online algorithms, such as ease of implementation and computational efficiency. Furthermore, we see the generality of our approach, which achieves strong performance across a number of diverse offline datasets, from those containing limited expert demonstrations, through to data comprised of high-volume sub-optimal trajectories.

We believe that our insights are valuable to the community and practitioners. We show that online off-policy RL algorithms can be remarkably effective at learning with offline data. However, we show their reliable performance is predicated on several key design choices, namely the way the offline data are sampled, a crucial way of normalizing the critic update, and using large ensembles to improve sample efficiency. While the individual ingredients of \ouralgo are refreshingly simple modifications on existing RL components, we show that their combination delivers state-of-the-art performance on a number of popular online RL with offline data benchmarks, exceeds the performance of significantly more complex prior methods, and generalizes to a number of different types of offline data, whether it be expert demonstrations or sub-optimal trajectories. We have released \ouralgo here: \href{https://github.com/ikostrikov/rlpd}{\texttt{github.com/ikostrikov/rlpd}}.
\section{Related work}

\paragraph{Offline RL pre-training.} We note connections to offline RL \cite{batchRL,bcq,offlineRL}; many prior works perform offline RL, followed by online fine-tuning \citep{dqfd, qt-opt, awac, off2on, iql}.
Notably, \citet{off2on} also considers large ensembles and multiple gradient-step per timestep regimes when learning online. However, our approach uses a significantly simpler sampling mechanism with no hyperparameters and does not rely on costly offline pre-training, which introduces yet \emph{additional} hyperparameters. We also emphasize that our normalized update is \emph{not an offline RL method}---we do not perform any offline pre-training but run online RL from scratch with offline data included in a replay buffer.

\paragraph{Constraining to prior data.} An alternative to the offline RL pre-training paradigm is to explicitly constrain the online agent updates such that it exhibits behavior that resembles the offline data \citep{gps, glearning, dqfd, ddpgBC, dapg, rudner2021on}. Particularly relevant to our approach is work by \citet{dapg}, which augments a policy gradient update with a weighted update that explicitly includes \emph{demonstration} data. In contrast, we use a sample-efficient off-policy paradigm, and \emph{do not} perform any pre-training. Also similar to our work is that by \citet{ddpgBC}, who also use an off-policy algorithm with a fixed offline replay buffer. However, we do not restrict the policy using a behavior cloning term, and do not reset to demonstration states. Moreover, we note that these approaches generally require the offline data to be high quality (i.e., `learning from demonstration data' \citep{playbackcontrol,LearningFromDemos}), while our approach is, importantly, agnostic to the quality of the data.

\paragraph{Unconstrained methods with prior data.} 
Prior work has also considered ways of incorporating offline data without any constraints. Some methods focus on initializing a replay buffer with offline data~\citep{ddpgfd,dqfd}, while other works have utilized a balanced sampling strategy to handle online and offline data~\citep{nair2018overcoming,qt-opt,modem,zhang2023policy}. Most recently, \citet{song2023hybrid} presented a theoretical analysis of such approaches, showing that balanced sampling is important both in theory and practice. In our experiments, we also show that balanced sampling helps online RL with offline data; however, directly using this approach on a range of benchmark tasks is insufficient, and other design decisions that we present are critical in achieving good performance across all tasks.
\section{Preliminaries}
\label{sec:preliminaries}
We consider problems that can be formulated as a Markov Decision Process (MDP) \citep{MDP}, described as a tuple $(\mathcal{S}, \mathcal{A}, \gamma, p, r, d_0)$ where $\mathcal{S}$ is the state space, $\mathcal{A}$ is the action space and $\gamma \in (0,1)$ is the discount factor. The dynamics are governed by a transition function $p(s'|s,a)$; there is a reward function $r(s,a)$ and initial state distribution $d_0(s)$. The goal of RL is then to maximize the expected sum of discounted rewards: $\mathbb{E}_\pi\left[ \sum_{t=1}^\infty \gamma^t r(s_t,a_t) \right].$ 

In this work, we focus on RL while having access to offline datasets $\mathcal{D}$ \cite{offlineRL}, a collection of $(s,a,r,s')$ tuples generated from a particular MDP. A key property of offline datasets is they usually do not provide complete state-action coverage, i.e., $\{s,a\in\mathcal{D}\}$ is a small subset of $\mathcal{S}\times\mathcal{A}$. Due to this lack of on-policy coverage, methods using function approximation may over-extrapolate values when learning on this data, leading to a pronounced effect on learning performance~\citep{bcq}.
\section{Online RL with Offline Data}
As outlined in~\cref{sec:preliminaries}, we consider the standard RL setting with the addition of a pre-collected dataset. In this work, we aim to design a general approach that is agnostic to the \emph{quality and quantity} of this pre-collected data. For instance, this data could take the form of a handful of human demonstrations, or swathes of sub-optimal, exploratory data. Furthermore, we wish to make recommendations that are agnostic to the nature of the problem setting, such as whether the observations are state or pixel-based, or whether the rewards are sparse or dense.

To this end, we present an approach based on off-policy model-free RL, without pre-training or explicit constraints, which we call \textbf{\ouralgo} (\textbf{R}einforcement \textbf{L}earning with \textbf{P}rior \textbf{D}ata).  As discussed in~\cref{sec:oreo}, we base our algorithm design on SAC~\citep{sac, sac2}, though in principle these design choices may improve other off-policy RL approaches. First, we propose a simple mechanism for incorporating the prior data. Then, we identify a pathology that exists when na\"ively applying off-policy methods to this problem setting, and propose a simple and minimally invasive solution. After, we improve the rate the offline data are utilized by incorporating the latest approaches in sample-efficient RL. Finally, we highlight common design choices in recent deep RL that are in fact environment sensitive, and should be adjusted accordingly by practitioners.

\subsection{Design Choice 1: A Simple and Efficient Strategy to Incorporate Offline Data}

We start with a simple approach that incorporates prior data which adds no computational overhead, yet is agnostic to the nature of the offline data. We call this `symmetric sampling', whereby for each batch we sample 50\% of the data from our replay buffer, and the \emph{remaining 50\% from the offline data buffer}, resembling the scheme used by \citet{rossbagnellagnostic}. As we will see in later sections, this sampling strategy is surprisingly effective across a variety of scenarios, and we extensively ablate various elements of this scheme (see \cref{sec:ablations}). However, applying this approach to canonical off-policy methods, such as SAC \citep{sac}, does not yield strong performance, as we see in \cref{fig:ourmethod}, and further design choices must be taken into consideration.

\subsection{Design Choice 2: Layer Normalization Mitigates Catastrophic Overestimation}

\label{sec:layernorm}

Standard off-policy RL algorithms query the learned Q-function for out-of-distribution (OOD) actions, which might not be defined during learning. Consequently, there can be significant overestimation of actual values due to the use of function approximation \citep{thrunOverestimate}. In practice, this phenomenon leads to training instabilities and possible \emph{divergence} when the critic is trying the catch up with a constantly increasing value.
\begin{figure}[t]
    \centering
    \vspace{-1mm}
    \includegraphics[width=0.49\textwidth]{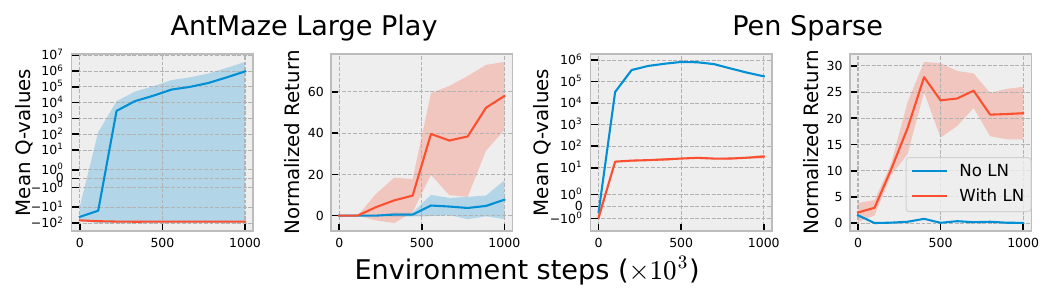} \\
    \vspace{-4.5mm}
    \caption{\small{Using SAC with our symmetric sampling method can result in instabilities due to diverging Q-values; with LayerNorm in the critic this disappears, improving performance.}}
    \label{fig:ln-sac}
    \vspace{-4mm}
\end{figure}

In particular, we find this to be the case when na\"ively applying our symmetric sampling approach for complex tasks (see~\cref{fig:ln-sac}).
Critic divergence is a well-studied problem, particularly in the \emph{offline} regime, where the policy cannot generate new experience. In our problem setting, however, we \emph{can} sample from the environment.
Therefore, instead of creating a mechanism that explicitly discourages OOD actions, which can be viewed as anti-exploration \citep{offlineRLAntiExplore}, we instead need to simply ensure that the learned functions do not extrapolate in an unconstrained manner. To this end, we show that Layer Normalization (LayerNorm) \citep{layernorm} can bound the extrapolation of networks but, crucially, \emph{does not} explicitly constrain the policy to remain close to the offline data. This in turn does not discourage the policy from exploring unknown and \emph{potentially valuable} regions of the state-action space. In particular, we demonstrate that LayerNorm bounds the values and empirically prevents catastrophic value extrapolation. Concretely, consider a Q-function $Q$ parameterized by $\theta, w$, applying LayerNorm and intermediate representation $\psi_\theta(\cdot,\cdot)$. For any $a$ and $s$ we can say\footnote{For simplicity, we consider LayerNorm without bias terms. This does not change the analysis, as it is a constant.}:
\begin{align*}
    \| Q_{\theta,w}(s,a) \| &= \|w^T \text{relu}(\psi_{\theta}(s,a)) \| \\
    &\le \|w\| \| \text{relu}(\psi_\theta(s,a))\| \le \|w\| \| \psi(s,a)\| \\
    &\le \|w\| 
\end{align*}
Therefore, as a result of \emph{Layer Normalization}, the Q-values are bounded by the norm of the weight layer, even for actions outside the dataset. Thus, the effect of erroneous action extrapolation is greatly mitigated, as their Q-values are unlikely to be significantly greater than those already seen in the data. Indeed, referring back to \cref{fig:ln-sac}, we see that introducing LayerNorm into the critic greatly improves performance through mitigating critic divergence.

\begin{figure}[t]
    \centering
    \vspace{1mm}
    \includegraphics[width=0.15\textwidth]{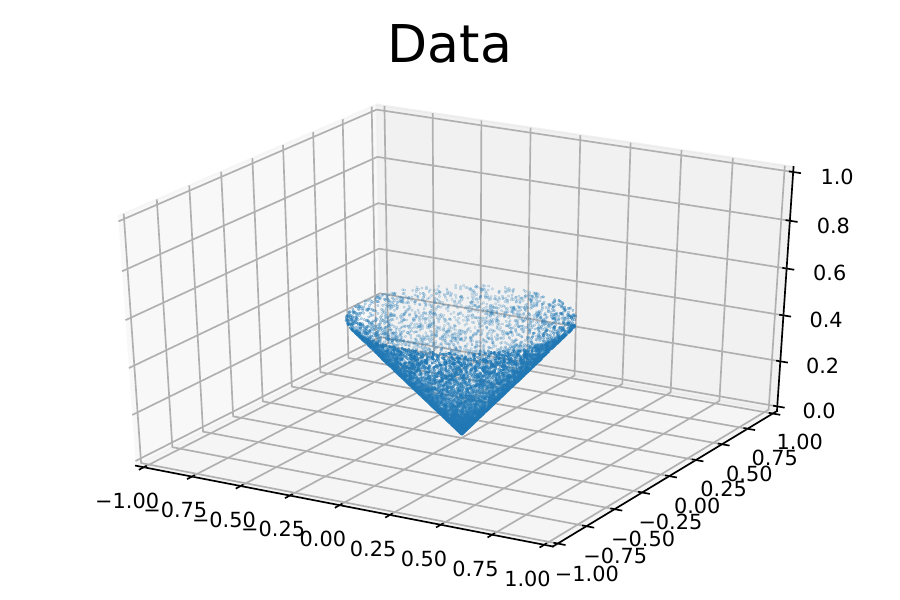}
    \includegraphics[width=0.15\textwidth]{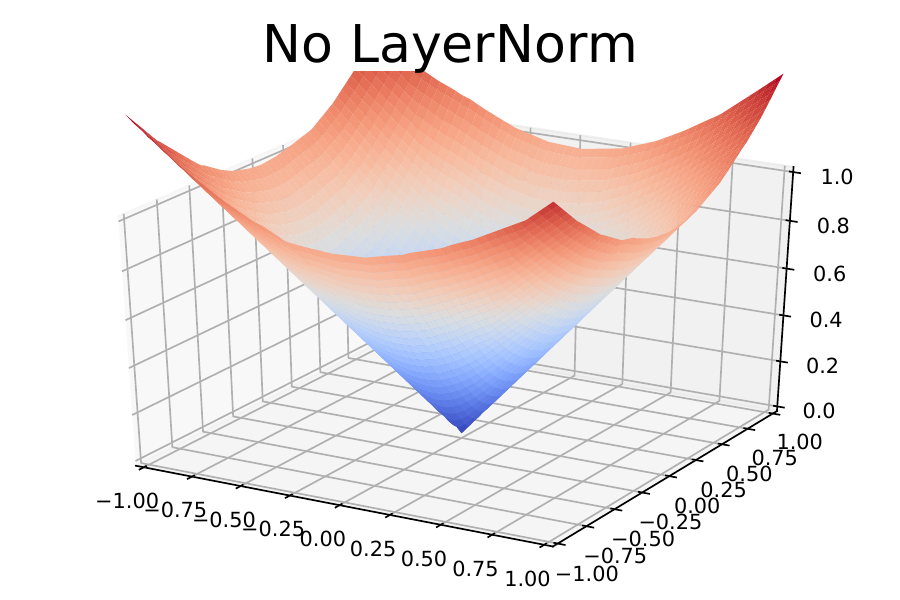}
    \includegraphics[width=0.15\textwidth]{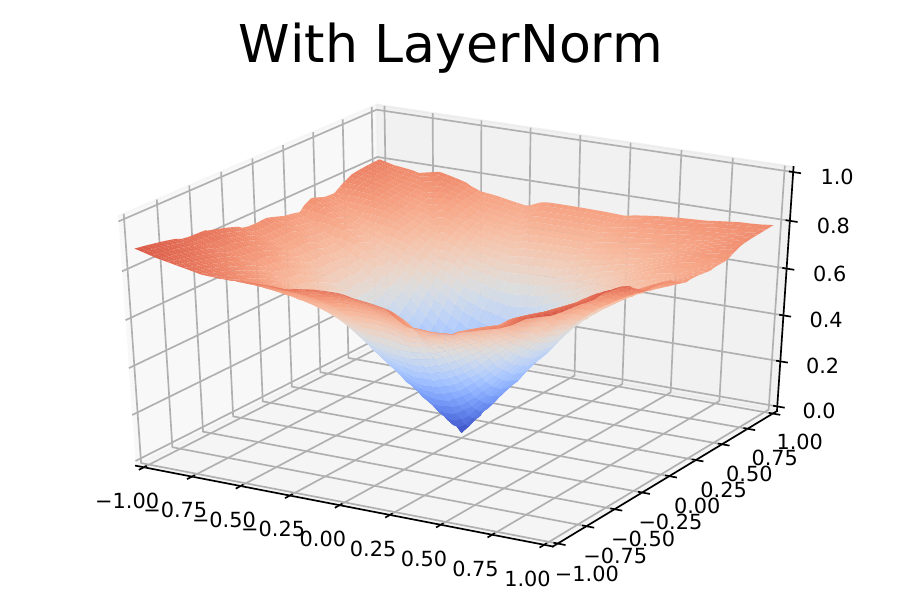}
    \vspace{-1mm}
    \caption{We fit data (left) with a two-layer MLP without LayerNorm (center) and with LayerNorm (right). LayerNorm bounds the values and prevents catastrophic overestimation.}
    \label{fig:toy_ln}
    \vspace{-4mm}
\end{figure}

To illustrate this, we generate a dataset with inputs $x$ distributed in a circle with radius $0.5$ and labels $y=\|x\|$. We study how a standard two-layer MLP with ReLU activations (common in deep RL) extrapolates outside of the data distribution, and the effect of adding LayerNorm. In \cref{fig:toy_ln}, the standard parameterization leads to unbounded extrapolation outside of the support, while LayerNorm bounds the values, greatly reducing the effect of uncontrolled extrapolation.

\subsection{Design Choice 3: Sample Efficient RL}

We now have an online approach leveraging offline data that also suppresses extreme value extrapolation, whilst maintaining the freedom of an unconstrained off-policy method. However, a benefit of offline and constrained approaches is that they have an explicit mechanism to efficiently incorporate prior data, such as through pre-training \citep{dqfd,off2on}, or an auxiliary supervision term \citep{ddpgBC,rudner2021on} respectively. In our case, the incorporation of prior data is implicit through the use of online Bellman backups over offline transitions. Therefore, it is imperative that these Bellman backups are performed as sample-efficiently as possible.

One way to achieve this is to increase the number of updates we perform per environment step (also referred to as update-to-data (UTD) ratio), allowing the offline data to become ``backed-up" more quickly.
However, as highlighted in recent literature in online RL, this can create issues in the optimization process and ironically \emph{reduce sample efficiency}, due to statistical over-fitting \citep{li2022efficient}. To ameliorate this, prior work has suggested a number of regularization approaches, such as simple L2 normalization \citep{ddpgfd}, Dropout \citep{dropout,droq} and random ensemble distillation \citep{redq}. In this work, we settle on the latter approach of random ensemble distillation; we will demonstrate through ablations that this performs strongest, particularly on sparse reward tasks. 

We also note that value over-fitting issues exist when performing TD-learning from images \citep{alix}. Therefore in these settings, we further include random shift augmentations \citep{drq, yarats2022mastering}.

\subsection{Per-Environment Design Choices}
\label{sec:perenv}

Having highlighted the 3 key design choices for our approach that can be applied generally to all environments and offline datasets, we now to turn our attention to design choices that are commonly taken for granted, but can in fact be environment-sensitive. It is well documented that deep RL algorithms are sensitive to implementation details \citep{deeprlthatmatters,Engstrom2020Implementation,andrychowicz2021what,furuta2021coadaptation}. As a result, many works in deep RL require per-environment hyperparameter tuning. Given the huge variety of tasks we consider in our experiments, we believe it is important to contribute to this discourse, and highlight that certain design choices, which are often simply inherited from previous implementations, should in fact be \emph{carefully reconsidered}, and may explain why off-policy methods have not been competitive thus far on the problems we consider. We therefore take a view that, given the well-documented sensitivity of deep RL, it is important to demonstrate a critical path of design choices to consider when assessing new environments, and provide a \emph{workflow} to simplify this process for practitioners.

\paragraph{Clipped Double Q-Learning (CDQ).}
\label{sec:doubleclipped}

Value-based methods combined with function approximation and stochastic optimization suffer from estimation uncertainty, which, in combination with the maximization objective of Q-learning, leads to value overestimation~\cite{Hasselt2015DeepRL} (as also discussed in \cref{sec:layernorm}).
In order to mitigate this issue, \citet{Fujimoto2018AddressingFA} introduce Clipped Double Q-Learning (CDQ) which involves taking a minimum of an ensemble of two Q-functions for computing TD-backups.
They define the targets for updating the critics as follows:
$$
y = r(s,a) + \gamma \min_{i=1,2} Q_{\theta_i}(s', a') \mbox{ where } a' \sim \pi(\cdot|s').
$$
However, this corresponds to fitting target Q-values that are 1 std.\ below the actual target values, and recent work \citep{top} suggests that this design choice may not be universally useful as it can be too conservative. Therefore, this is important to reconsider, especially outside of the domains for which it was originally designed, such as sparse reward tasks prevalent in our problem setting.

\paragraph{Maximum Entropy RL.}
MaxEnt RL augments the traditional return objective with an entropy term:
$$
\max_{\pi} \mathbb{E}_{s \sim \rho^\pi, a \sim \pi} \left[ \sum_{t=0}^{\infty} \gamma^t (r_t + \alpha \mathcal{H}(\pi(a|s))) \right].
$$

This corresponds to maximizing the discounted reward and expected policy entropy at each time step. The motivation for these approaches is centered around robustness and exploration (i.e., maximize reward while behaving as randomly as possible). Approaches relying on this objective are empirically impressive~\citep{sac,sac2,redq,droq}. We, therefore, believe this design choice is of interest in the context of online fine-tuning, where rewards are often sparse and require exploration.

\paragraph{Architecture.}

Network architecture can have a significant impact on deep RL performance, and the same architecture that can be optimal in one environment can be sub-optimal in another \cite{furuta2021coadaptation}. To simplify the search space, we consider the impact of having 2 or 3 layers in the actor and critic, which have been shown to affect performance, even on canonical tasks \citep{offcon3}.

\subsection{\ouralgo: Approach Overview}
\label{sec:oreo}

Here we present pseudo-code for our approach, highlighting in \textcolor{ForestGreen}{Green} elements that are important to our approach, and in \textcolor{Orchid}{Purple}, environment-specific design choices.

The key factors of \ouralgo reside in lines 1 and 13 of \cref{algo:simple} with adopting LayerNorm, large ensembles, sample efficient learning and a symmetric sampling approach to incorporate online and offline data. For environment specific choices, we recommend the following as a starting point: 
\begin{itemize}
\vspace{-2.5mm}
\setlength\itemsep{0.05em}
\item {\em Line 3:} Subset 2 critics 
\item {\em Line 16:} Remove entropy
\item {\em Line 1:} Utilize a \emph{deeper} 3 layer MLP
\vspace{-2.5mm}
\end{itemize}
As a pragmatic workflow, we recommend ablating these \textcolor{Orchid}{Purple} design choices first, and in the order stated above.
\begin{algorithm}[!t]
  \caption{Online RL with Offline Data (\ouralgo) \label{algo:simple}}
  \begin{algorithmic}[1]
    \STATE \mathcolor{ForestGreen}{Select LayerNorm, Large Ensemble Size $E$, Gradient Steps $G$}\mathcolor{Orchid}{, and architecture.}
    \STATE Randomly initialize Critic $\theta_i$ (set targets $\theta_i' = \theta_i$) for $i = 1,2,\dots, E$ and Actor $\phi$ parameters. Select discount $\gamma$, temperature $\alpha$ and critic EMA weight $\rho$.
    \STATE \mathcolor{Orchid}{Determine number of Critic targets to subset $Z\in\{1,2\}$}
    \STATE Initialize empty replay buffer $\mathcal{R}$
    \STATE \mathcolor{ForestGreen}{Initialize buffer $\mathcal{D}$ with offline data}
    \WHILE{True}
      \STATE Receive initial observation state $s_0$
      \FOR{t = 0, T}
        \STATE Take action $a_t \sim \pi_\phi(\cdot|s_t)$
        \STATE Store transition $(s_t, a_t,
                r_t, s_{t+1})$ in $\mathcal{R}$
        \FOR{\mathcolor{ForestGreen}{$g = 1, G$}}
            \STATE Sample minibatch $b_R$ of \mathcolor{ForestGreen}{$\frac{N}{2}$} 
            from $\mathcal{R}$
            \STATE \mathcolor{ForestGreen}{Sample minibatch $b_D$ 
            of $\frac{N}{2}$ 
            from $\mathcal{D}$}
            \STATE \mathcolor{ForestGreen}{Combine $b_R$ and $b_D$ to form batch $b$ of size $N$}
            \STATE Sample set $\mathcal{Z}$ of $Z$ indices from $\{1, 2, \dots, E\}$
            \STATE With $b$, set
            \vspace{-2mm}
            \[ y = r + \gamma \left( \min_{i\in\mathcal{Z}}Q_{\theta_i'}(s',
            \tilde{a}')\right), \quad \tilde{a}'\sim\pi_\phi(\cdot|s')\]
            \vspace{-2mm}
            \STATE \mathcolor{Orchid}{Add entropy term $ y = y + \gamma\alpha \log \pi_\phi(\tilde{a}'|s')$}
            \FOR{$i=1, E$}
                \STATE Update $\theta_i$ minimizing loss:
                    \vspace{-2mm}
                       \[L = \frac{1}{N} \sum_i (y -
                       Q_{\theta_i}(s, a))^2\]
                    \vspace{-4mm}
            \ENDFOR
            \STATE Update target networks $\theta_i' \leftarrow \rho\theta_i' + (1-\rho)\theta_i$
        \ENDFOR
        \STATE With $b$, update $\phi$ maximizing objective:
        \vspace{-2mm}
        \[
            \frac{1}{E}\sum_{i =1}^{E} Q_{\theta_{i}}(s, \tilde{a})-\alpha \log \pi_{\phi}(\tilde{a} | s), \quad
 \tilde{a} \sim \pi_{\phi}(\cdot \mid s)
         \]
        \vspace{-3mm}
        \ENDFOR
    \ENDWHILE
  \end{algorithmic}
\end{algorithm}

\begin{figure*}[t]
    \centering
    \vspace{-2.75mm}
    \includegraphics[width=0.98\textwidth]{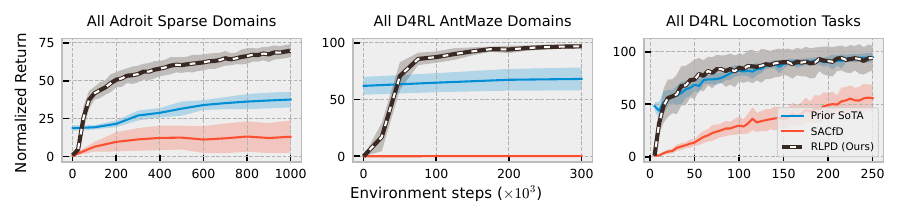} \\
    \vspace{-7.25mm}
    \caption{\ouralgo exceeds prior state-of-the-art performance on a number of different popular benchmarks whilst being significantly simpler. Results are aggregated over \textbf{21 different environments} (10 Seeds, 1 std. shaded). In each case, we compare to the prior best known work (IQL + Finetuning in Adroit and AntMaze, Off2On in Locomotion), and SACfD, a canonical off-policy approach using offline data.} 
    \label{fig:all_results}
    \vspace{-4mm}
\end{figure*}

\section{Experiments}

We design our experiments to not only demonstrate the importance of our design choices, but also provide the insights that allow practitioners to quickly adapt \ouralgo to their problems. As such, we aim to answer the following questions:

\begin{enumerate}[topsep=0.01em,leftmargin=*]
    \item Is \ouralgo competitive with prior work despite using \emph{no pre-training nor having explicit constraints}?
    \item Does \ouralgo transfer to \emph{pixel-based} environments?
    \item Does LayerNorm \emph{mitigate value divergence}?
    \item Does the proposed workflow around environment-specific design choices lead to \emph{reliable performance}?
\end{enumerate}

For (1), we compare \ouralgo to those which have been designed to use offline data to accelerate online learning. For (2), we consider an additional suite of tasks to study \ouralgo's applicability to vision-based domains. 
Then, for (3) we perform analysis to demonstrate the importance of using LayerNorm. Lastly, for (4) we demonstrate our proposed workflow on the most challenging tasks (see Subsection~\ref{sec:oreo}).

\paragraph{How does \ouralgo compare?} We consider the following \textbf{21 tasks} from established benchmarks:
\begin{itemize}[nosep,leftmargin=*]
    \item \textbf{Sparse Adroit}~\citep{awac}. These 3 dexterous manipulation tasks---pen-spinning, door-opening, ball relocation---are challenging, sparse-reward tasks. The offline data are multi-modal, with a small set of human demonstrations and a large set of trajectories from a behavior-cloned policy trained on this human data.
    We follow the \emph{rigorous evaluation criteria} of~\citet{iql}, whereby performance is based on completion speed, rather than success rate. IQL + Finetuning represents the strongest prior work~\citep{iql}.
    \item \textbf{D4RL AntMaze} \citep{fu2020d4rl}. These 6 sparse reward tasks require an Ant agent to learn to walk and navigate through a maze to reach a goal. The offline data comprise \emph{only sub-optimal} trajectories that can in principle be ``stitched" together. Again, IQL + Fine-tuning represents the strongest prior work \citep{iql}.
    \item \textbf{D4RL Locomotion}~\citep{fu2020d4rl}. Lastly, we have 12 dense reward, locomotion tasks featuring offline data with varying levels of expertise. Off2On~\citep{off2on} has state-of-the-art performance on this suite of tasks.
\end{itemize}

For evaluation, we first include \textbf{SACfD}, a baseline studied in prior work~\citep{ddpgfd,awac}, which, similar to \ouralgo, is an off-policy approach that incorporates offline data during training. However, SACfD simply \emph{initializes} the online replay buffer with the offline data.
We implement this baseline using SAC without the additional design decisions discussed in~\autoref{sec:oreo}. Then, since there is no \emph{single} prior method that achieves the best performance across all groups of tasks, we compare to the state-of-the-art method specific to each group (as listed above).We refer to this comparison in our plots as \textbf{Prior SoTA}. For all experiments in this work, we report the mean and standard deviation across \emph{10 seeds} and aggregate the results across tasks within the three groups listed. For full detailed results broken down by task, see \cref{sec:detailedexperiments}, and for more environment details, see \cref{sec:envdetails}.

We see that \ouralgo performs strongly, either matching or \emph{significantly exceeding} the best known prior work on these challenging benchmarks(see~\cref{fig:all_results}). We reiterate that we present results for \ouralgo and SACfD without doing \emph{any pre-training}, unlike the Prior SoTA methods. So, while prior work (shown in blue) may achieve strong initial performance, the online improvement is more modest. On the other hand, our method reaches or surpasses this performance in the order of just 10k online samples.
Notably, we outperform the best reported performance on the Sparse Adroit `Door' task by $\mathbf{2.5\times}$. Moreover, to our knowledge, our method is the first to effectively `solve' \emph{all} AntMaze tasks. Furthermore, we are able to do so in \emph{less than a third} the time-step budget allocated to prior methods.%

\paragraph{Does \ouralgo transfer to pixels?} Here we consider the medium and expert locomotion tasks in V-D4RL \cite{lu2022challenges}, an offline dataset with \emph{only pixel observations}. V-D4RL is particularly challenging as the behavior policies are state-based, which means the data is partially observable in pixel-space. This is most obvious in `Humanoid Walk', where body parts are visually occluded, thus behavior cloning (BC) can struggle to achieve strong performance. For evaluation, as our method seeks to accelerate online learning with offline data, we focus on data-efficiency---we introduce a challenge that we call ``10\%DMC", which involves training policies using only 10\% of the total timesteps recommended by \citet{yarats2022mastering}. As we use the medium and expert data, we include a \textbf{BC} baseline. To evaluate \ouralgo's ability to efficiently use offline data to boost online learning, we compare to a baseline approach that does not use the offline data. To isolate the effect of the utilization of offline data, we use the same architecture and policy optimizer as our method and label this baseline as \textbf{Online} in our plots. Then, to evaluate how this compares to the state-of-the-art sample-efficient RL method from pixels, we compare to~\textbf{DrQ-v2} \citep{yarats2022mastering}. 

\begin{figure}[H]
    \centering
    \vspace{-3mm}
    \includegraphics[width=0.49\textwidth]{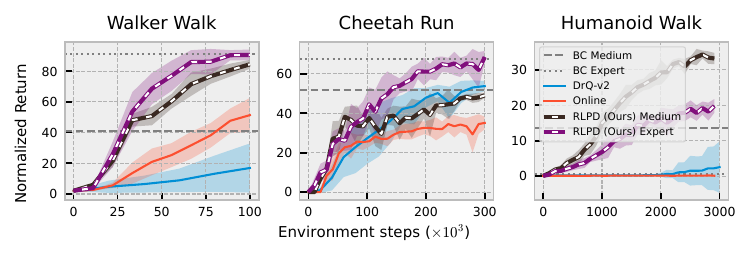} \\
    \vspace{-5mm}
    \caption{Our approach generalizes to vision-based domains, providing consistent improvements over existing approaches.}
    \label{fig:vd4rl}
    \vspace{-5mm}
\end{figure}

We see in~\cref{fig:vd4rl}\footnote{V-D4RL normalized return is episode return divided by 10.} that \ouralgo provides consistent improvements over purely online approaches, and in many cases greatly improves over a BC baseline. We see through the difference in \ouralgo (dashed black and purple lines) and the Online baseline that \ouralgo \emph{effectively} utilizes the offline data to bootstrap learning. This conclusion holds as we compare to the SoTA vision-based RL method (blue).
\begin{wrapfigure}{R}{0.2\textwidth}
    \centering
    \vspace{-6mm}
    \includegraphics[width=0.2\textwidth,trim={2mm 0 1mm 0}, clip]{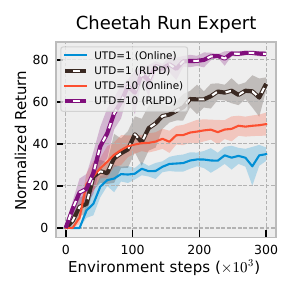} \\
    \vspace{-6mm}
    \caption{Increasing UTD with \ouralgo greatly improves sample efficiency from pixels.}
    \label{fig:vd4rl-highutd}
    \vspace{-4mm}
\end{wrapfigure}

Lastly, we test increasing UTD to 10 on one of the tasks---Cheetah Run with Expert offline data. \cref{fig:vd4rl-highutd} shows a remarkable improvement in performance when learning with the offline dataset. To our knowledge, this is the first demonstration of a high UTD approach improving model-free pixel-based continuous control.

\subsection{\ouralgo Analysis and Ablation Study}
\label{sec:ablations}
Here, we address (3) and (4) by quantifying the effect of LayerNorm, and demonstrating the reliability of our proposed workflow (see Subsection~\ref{sec:oreo}).
\paragraph{Does LayerNorm mitigate value divergence?}

\begin{figure}[t]
    \vspace{-3mm}
    \centering
    \includegraphics[width=0.48\textwidth]{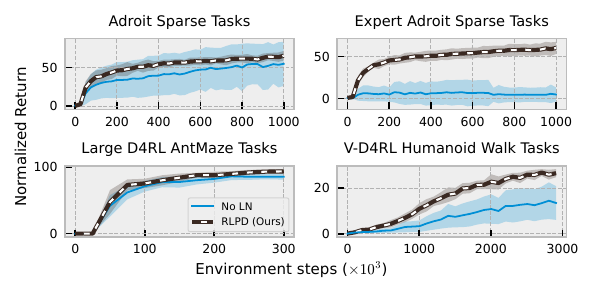} \\
    \vspace{-6mm}
    \caption{LayerNorm is crucial for strong performance, particularly when data are limited or narrowly distributed.}
    \label{fig:layernormablate}
    \vspace{-5mm}
\end{figure}

We now illustrate the importance of LayerNorm in mitigating catastrophic value divergence, and focus on tasks where overestimation is particularly prone. 
In \cref{fig:layernormablate}, we see LayerNorm is crucial for strong performance in the Adroit domain; excluding LayerNorm results in significantly higher variance across seeds and reduces mean performance.

To more clearly illustrate this effect, we construct a dataset of \emph{only the expert human demonstration data} from the Adroit Sparse tasks (see ``Expert Adroit Sparse Tasks" in \cref{fig:layernormablate}). This subset comprises just \textbf{22} of the 500 trajectories in the original dataset and is much more narrowly distributed by nature---representing a task with sparse rewards, limited demonstrations, and narrow offline data coverage---likely to exacerbate value divergence.
Here we see a remarkable result: \ouralgo still \emph{exceeds} prior work, despite significant restrictions in data. Moreover, removing LayerNorm now results in collapsed performance, with no progress made on any task. We further observe improvements in sample efficiency through the inclusion of LayerNorm in AntMaze and Humanoid Walk through reducing excessive extrapolation. Additional experiments and results are in \cref{sec:detailedexperiments}.

\paragraph{Design choice workflow.}
We now motivate our workflow in Section~\ref{sec:oreo}, demonstrating the importance of these design choices. We focus on the hardest tasks, namely `Relocate' in Adroit Sparse, `Large Diverse' in AntMaze, and `Humanoid Expert' in V-D4RL, as we found these to be most sensitive. We provide full results on all tasks in \cref{sec:detailedexperiments}, noting that optimal decisions in the harder domains also produce optimal results in the easier domains.

\begin{figure}[H]
    \centering
    \vspace{-3mm}
    \includegraphics[width=0.49\textwidth]{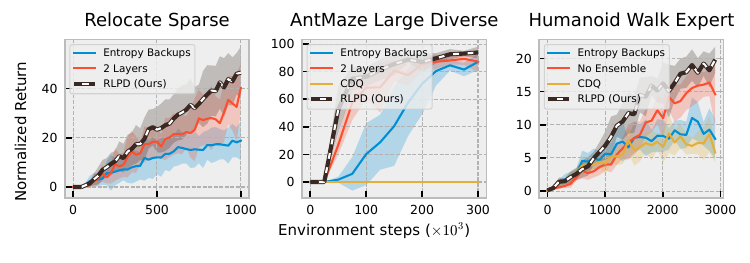} \\
    \vspace{-6mm}
    \caption{Our recommended starting design choices and workflow leads to strong performance on all tasks.}
    \label{fig:workflow}
    \vspace{-4mm}
\end{figure}

We see in \cref{fig:workflow} that with the recommended environment-specific design choices provide strong performance; we see using entropy backups and smaller networks always results in worse performance. In `AntMaze Large Diverse' however we see that with CDQ, performance deteriorates. Following our workflow, and ablating this by subsetting 1 critic is crucial to recovering strong performance. The same applies to `Humanoid Walk Expert', whereby we surprisingly see that CDQ is detrimental to performance, despite its popularity in recent implementations. We also show the surprising positive effect of larger ensembles in pixel-based tasks, with the standard 2 member critic ensemble performing worse than the 10 member ensemble we use by default in \ouralgo.

Lastly, we conduct additional ablations to understand the importance of the design choices that we propose for our method, showing that, though the individual design decisions are simple, they are vital for good performance.

\paragraph{Critic regularization.}
Here we examine the effects of critic regularization on performance. We compare 3 approaches: weight-decay\footnote{We found a weight decay value of 0.01 worked best across a number of settings.} \citep{ddpgfd}, Dropout \citep{droq} and ensembling \citep{redq}. We choose a subset of the prior experiments to evaluate on from the AntMaze, Adroit, and Locomotion sections.

\begin{figure}[H]
    \centering
    \vspace{-2mm}
    \includegraphics[width=0.49\textwidth]{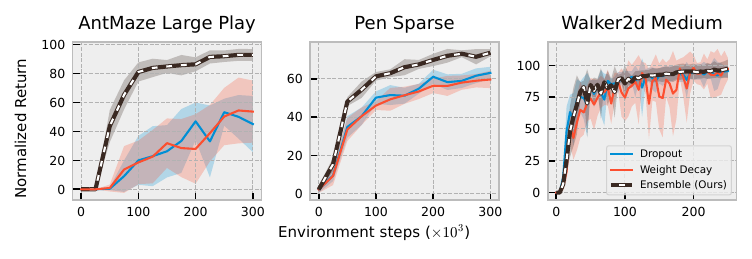} \\
    \vspace{-5mm}
    \caption{In general, critic ensembling provides the best performance. Dropout performs worse in sparse reward tasks.}
    \label{fig:reg}
    \vspace{-3mm}
\end{figure}

In \cref{fig:reg}, we see that ensembling is the strongest form of regularization. Notably, while Dropout performs well in the Locomotion domain `walker2d-medium-v0', as affirmed by \citet{droq}, it does not generalize to challenging sparse reward environments. We also see that weight-decay regularization is less performant in all domains.

\paragraph{Buffer initialization.}
In this section, we compare symmetric sampling with that of one which relies on \emph{initializing} a replay buffer with offline data.

\begin{figure}[H]
    \centering
    \vspace{-2mm}
    \includegraphics[width=0.45\textwidth]{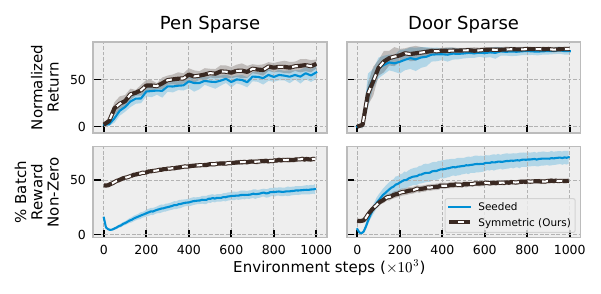} \\
    \vspace{-5mm}
    \caption{Symmetric sampling improves sample efficiency and reduces variance across seeds, and does not work by simply increasing the reward density in a batch.}
    \label{fig:buffer-adroit}
    \vspace{-3mm}
\end{figure}
\begin{wrapfigure}{R}{0.2\textwidth}
    \centering
    \vspace{-5mm}
    \includegraphics[width=0.2\textwidth,trim={2mm 0 1mm 0}, clip]{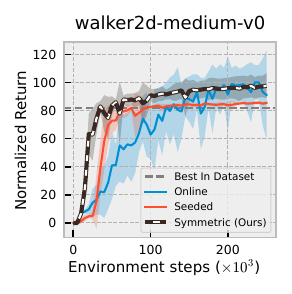} \\
    \vspace{-4mm}
    \caption{Initializing the buffer with large amounts of data limits improvement.}
    \label{fig:buffer-loco}
    \vspace{-3mm}
\end{wrapfigure}
First returning to the challenging Human Expert Demonstrations setting in Adroit, in \cref{fig:buffer-adroit} we show two contrasting examples that demonstrate symmetric sampling effectively trades-off between replay and offline data. We observe that in the Pen environment, symmetric sampling clearly improves exploration by ensuring increasing reward density within mini-batches. In contrast, we see that although the buffer initialization approach explores just as well in the Door environment, symmetric sampling has the important effect of improving stability and decreases variance due to relying less on the higher-variance data generated by the online policy.

We now consider a situation whereby we have abundant sub-optimal offline data, and see that again, our balanced sampling approach improves sample-efficiency. In \cref{fig:buffer-loco}, initializing the buffer with high volumes of medium quality locomotion data gives initial improvement over online performance, but struggles asymptotically, likely due to a lack of on-policy data sampling, vital for online improvement. On the other hand, our symmetric sampling approach improves sample efficiency and matches asymptotic performance while reducing variance.

\paragraph{Sampling proportion sensitivity.}

We assess the sensitivity of our sampling approach away away from the ``symmetric" ratio of 50\% online/offline.
\begin{figure}[h]
    \centering
    \vspace{-1mm}
    \includegraphics[width=0.49\textwidth]{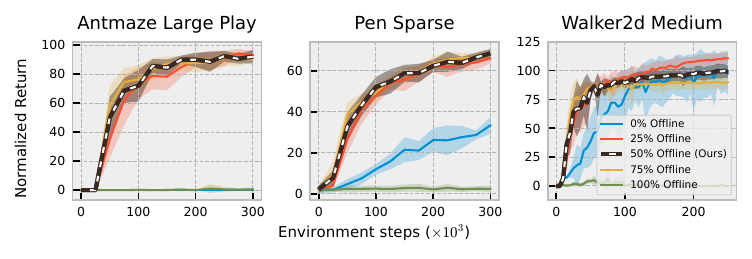} \\
    \vspace{-5mm}
    \caption{\ouralgo is not sensitive to replay proportion; 50\% offers the best compromise between variance, speed of convergence, and asymptotic performance.}
    \label{fig:prop}
    \vspace{-3mm}
\end{figure}

As we see in \cref{fig:prop}, \ouralgo is not very sensitive to sampling proportion. While sampling 25\% offline can marginally help asymptotic performance in `walker2d-medium-v0', this comes at the expense of variance and sample efficiency in sparse reward tasks. We also affirm with our 100\% offline results that \ouralgo is not an offline method, and key to its success is how it controls for divergence \emph{without restricting exploration or behavior learning}.
\section{Conclusion}

In this work, we show that off-policy approaches can be adapted to leverage offline data when training online.
We see that with careful application of key design choices, \ouralgo can attain remarkably strong performance, and demonstrate this on a total of \textbf{30 different tasks}. Concretely, we show that the unique combination of symmetric sampling, LayerNorm as a value extrapolation regularizer, and sample efficient learning is key to its success, resulting in our outperforming prior work by up to $\mathbf{2.5\times}$ on a large variety of competitive benchmarks. Moreover, our recommendations have negligible impact on computational efficiency compared to pure-online approaches, and are simple, allowing practitioners to easily incorporate the insights of this work into existing approaches. To further improve adoptability, we recommend and demonstrate a workflow for practitioners that improves performance on a wide variety of tasks, and thus demonstrate that certain canonical design choices should be reconsidered when applying off-policy methods. Finally, to facilitate future research, we have released the \ouralgo codebase here: \href{https://github.com/ikostrikov/rlpd}{\texttt{github.com/ikostrikov/rlpd}}, which features highly optimized off-policy algorithms for proprioceptive and pixel-based tasks in a single codebase written in JAX \citep{jax2018github}.
\section*{Acknowledgements}
PJB is funded through the Willowgrove Foundation and the Les Woods Memorial Fund. This research was partly supported by the DARPA RACER program, ARO W911NF-21-1-0097, the Office of Naval Research under N00014-21-1-2838 and N00014-19-12042. This research used the Savio computational cluster resource provided by the Berkeley Research Computing program at the University of California, Berkeley (supported by the UC Berkeley Chancellor, Vice Chancellor for Research, and Chief Information Officer).

\bibliography{paper}

\begin{thebibliography}{52}
\providecommand{\natexlab}[1]{#1}
\providecommand{\url}[1]{\texttt{#1}}
\expandafter\ifx\csname urlstyle\endcsname\relax
  \providecommand{\doi}[1]{doi: #1}\else
  \providecommand{\doi}{doi: \begingroup \urlstyle{rm}\Url}\fi

\bibitem[Andrychowicz et~al.(2021)Andrychowicz, Raichuk, Sta{\'n}czyk, Orsini,
  Girgin, Marinier, Hussenot, Geist, Pietquin, Michalski, Gelly, and
  Bachem]{andrychowicz2021what}
Andrychowicz, M., Raichuk, A., Sta{\'n}czyk, P., Orsini, M., Girgin, S.,
  Marinier, R., Hussenot, L., Geist, M., Pietquin, O., Michalski, M., Gelly,
  S., and Bachem, O.
\newblock What matters for on-policy deep actor-critic methods? a large-scale
  study.
\newblock In \emph{International Conference on Learning Representations}, 2021.
\newblock URL \url{https://openreview.net/forum?id=nIAxjsniDzg}.

\bibitem[Asada \& Hanafusa(1979)Asada and Hanafusa]{playbackcontrol}
Asada, H. and Hanafusa, H.
\newblock Playback control of force teachable robots.
\newblock \emph{Transactions of the Society of Instrument and Control
  Engineers}, 15\penalty0 (3):\penalty0 410--411, 1979.
\newblock \doi{10.9746/sicetr1965.15.410}.

\bibitem[Ba et~al.(2016)Ba, Kiros, and Hinton]{layernorm}
Ba, J.~L., Kiros, J.~R., and Hinton, G.~E.
\newblock Layer normalization, 2016.
\newblock URL \url{https://arxiv.org/abs/1607.06450}.

\bibitem[Ball \& Roberts(2021)Ball and Roberts]{offcon3}
Ball, P.~J. and Roberts, S.~J.
\newblock Offcon$^3$: What is state of the art anyway?, 2021.
\newblock URL \url{https://arxiv.org/abs/2101.11331}.

\bibitem[Bellman(1957)]{MDP}
Bellman, R.
\newblock A markovian decision process.
\newblock \emph{Indiana Univ. Math. J.}, 6:\penalty0 679--684, 1957.
\newblock ISSN 0022-2518.

\bibitem[Bradbury et~al.(2018)Bradbury, Frostig, Hawkins, Johnson, Leary,
  Maclaurin, Necula, Paszke, Vander{P}las, Wanderman-{M}ilne, and
  Zhang]{jax2018github}
Bradbury, J., Frostig, R., Hawkins, P., Johnson, M.~J., Leary, C., Maclaurin,
  D., Necula, G., Paszke, A., Vander{P}las, J., Wanderman-{M}ilne, S., and
  Zhang, Q.
\newblock {JAX}: composable transformations of {P}ython+{N}um{P}y programs,
  2018.
\newblock URL \url{http://github.com/google/jax}.

\bibitem[Cabi et~al.(2019)Cabi, Colmenarejo, Novikov, Konyushkova, Reed, Jeong,
  Zolna, Aytar, Budden, Vecer{\'i}k, Sushkov, Barker, Scholz, Denil,
  de~Freitas, and Wang]{Cabi2019ScalingDR}
Cabi, S., Colmenarejo, S.~G., Novikov, A., Konyushkova, K., Reed, S.~E., Jeong,
  R., Zolna, K., Aytar, Y., Budden, D., Vecer{\'i}k, M., Sushkov, O.~O.,
  Barker, D., Scholz, J., Denil, M., de~Freitas, N., and Wang, Z.
\newblock Scaling data-driven robotics with reward sketching and batch
  reinforcement learning.
\newblock \emph{Robotics: Science and Systems XVI}, 2019.

\bibitem[Cetin et~al.(2022)Cetin, Ball, Roberts, and Celiktutan]{alix}
Cetin, E., Ball, P.~J., Roberts, S., and Celiktutan, O.
\newblock Stabilizing off-policy deep reinforcement learning from pixels.
\newblock In Chaudhuri, K., Jegelka, S., Song, L., Szepesvari, C., Niu, G., and
  Sabato, S. (eds.), \emph{Proceedings of the 39th International Conference on
  Machine Learning}, volume 162 of \emph{Proceedings of Machine Learning
  Research}, pp.\  2784--2810. PMLR, 17--23 Jul 2022.
\newblock URL \url{https://proceedings.mlr.press/v162/cetin22a.html}.

\bibitem[Chen et~al.(2021)Chen, Wang, Zhou, and Ross]{redq}
Chen, X., Wang, C., Zhou, Z., and Ross, K.~W.
\newblock Randomized ensembled double q-learning: Learning fast without a
  model.
\newblock In \emph{International Conference on Learning Representations}, 2021.
\newblock URL \url{https://openreview.net/forum?id=AY8zfZm0tDd}.

\bibitem[Engstrom et~al.(2020)Engstrom, Ilyas, Santurkar, Tsipras, Janoos,
  Rudolph, and Madry]{Engstrom2020Implementation}
Engstrom, L., Ilyas, A., Santurkar, S., Tsipras, D., Janoos, F., Rudolph, L.,
  and Madry, A.
\newblock Implementation matters in deep rl: A case study on ppo and trpo.
\newblock In \emph{International Conference on Learning Representations}, 2020.
\newblock URL \url{https://openreview.net/forum?id=r1etN1rtPB}.

\bibitem[Ernst et~al.(2005)Ernst, Geurts, and Wehenkel]{batchRL}
Ernst, D., Geurts, P., and Wehenkel, L.
\newblock Tree-based batch mode reinforcement learning.
\newblock \emph{Journal of Machine Learning Research}, 6\penalty0
  (18):\penalty0 503--556, 2005.
\newblock URL \url{http://jmlr.org/papers/v6/ernst05a.html}.

\bibitem[Fox et~al.(2016)Fox, Pakman, and Tishby]{glearning}
Fox, R., Pakman, A., and Tishby, N.
\newblock Taming the noise in reinforcement learning via soft updates.
\newblock In \emph{32nd Conference on Uncertainty in Artificial Intelligence
  (UAI)}, 2016.

\bibitem[Fu et~al.(2020)Fu, Kumar, Nachum, Tucker, and Levine]{fu2020d4rl}
Fu, J., Kumar, A., Nachum, O., Tucker, G., and Levine, S.
\newblock D4rl: Datasets for deep data-driven reinforcement learning, 2020.

\bibitem[Fujimoto et~al.(2018)Fujimoto, van Hoof, and
  Meger]{Fujimoto2018AddressingFA}
Fujimoto, S., van Hoof, H., and Meger, D.
\newblock Addressing function approximation error in actor-critic methods.
\newblock \emph{ArXiv}, abs/1802.09477, 2018.

\bibitem[Fujimoto et~al.(2019)Fujimoto, Meger, and Precup]{bcq}
Fujimoto, S., Meger, D., and Precup, D.
\newblock Off-policy deep reinforcement learning without exploration.
\newblock In Chaudhuri, K. and Salakhutdinov, R. (eds.), \emph{Proceedings of
  the 36th International Conference on Machine Learning}, volume~97 of
  \emph{Proceedings of Machine Learning Research}, pp.\  2052--2062. PMLR,
  09--15 Jun 2019.
\newblock URL \url{https://proceedings.mlr.press/v97/fujimoto19a.html}.

\bibitem[Furuta et~al.(2021)Furuta, Kozuno, Matsushima, Matsuo, and
  Gu]{furuta2021coadaptation}
Furuta, H., Kozuno, T., Matsushima, T., Matsuo, Y., and Gu, S.
\newblock Co-adaptation of algorithmic and implementational innovations in
  inference-based deep reinforcement learning.
\newblock In Beygelzimer, A., Dauphin, Y., Liang, P., and Vaughan, J.~W.
  (eds.), \emph{Advances in Neural Information Processing Systems}, 2021.
\newblock URL \url{https://openreview.net/forum?id=vLyI__SoeAe}.

\bibitem[Haarnoja et~al.(2018{\natexlab{a}})Haarnoja, Zhou, Abbeel, and
  Levine]{sac}
Haarnoja, T., Zhou, A., Abbeel, P., and Levine, S.
\newblock Soft actor-critic: Off-policy maximum entropy deep reinforcement
  learning with a stochastic actor.
\newblock \emph{International Conference on Machine Learning (ICML)},
  2018{\natexlab{a}}.

\bibitem[Haarnoja et~al.(2018{\natexlab{b}})Haarnoja, Zhou, Hartikainen,
  Tucker, Ha, Tan, Kumar, Zhu, Gupta, Abbeel, and Levine]{sac2}
Haarnoja, T., Zhou, A., Hartikainen, K., Tucker, G., Ha, S., Tan, J., Kumar,
  V., Zhu, H., Gupta, A., Abbeel, P., and Levine, S.
\newblock Soft actor-critic algorithms and applications, 2018{\natexlab{b}}.
\newblock URL \url{https://arxiv.org/abs/1812.05905}.

\bibitem[Hansen et~al.(2022)Hansen, Lin, Su, Wang, Kumar, and
  Rajeswaran]{modem}
Hansen, N., Lin, Y., Su, H., Wang, X., Kumar, V., and Rajeswaran, A.
\newblock Modem: Accelerating visual model-based reinforcement learning with
  demonstrations.
\newblock \emph{arXiv preprint}, 2022.

\bibitem[Henderson et~al.(2018)Henderson, Islam, Bachman, Pineau, Precup, and
  Meger]{deeprlthatmatters}
Henderson, P., Islam, R., Bachman, P., Pineau, J., Precup, D., and Meger, D.
\newblock Deep reinforcement learning that matters.
\newblock \emph{Proceedings of the AAAI Conference on Artificial Intelligence},
  32\penalty0 (1), Apr. 2018.
\newblock \doi{10.1609/aaai.v32i1.11694}.
\newblock URL \url{https://ojs.aaai.org/index.php/AAAI/article/view/11694}.

\bibitem[Hester et~al.(2018)Hester, Vecerik, Pietquin, Lanctot, Schaul, Piot,
  Horgan, Quan, Sendonaris, Osband, Dulac-Arnold, Agapiou, Leibo, and
  Gruslys]{dqfd}
Hester, T., Vecerik, M., Pietquin, O., Lanctot, M., Schaul, T., Piot, B.,
  Horgan, D., Quan, J., Sendonaris, A., Osband, I., Dulac-Arnold, G., Agapiou,
  J., Leibo, J., and Gruslys, A.
\newblock Deep q-learning from demonstrations.
\newblock \emph{Proceedings of the AAAI Conference on Artificial Intelligence},
  32\penalty0 (1), Apr. 2018.
\newblock \doi{10.1609/aaai.v32i1.11757}.
\newblock URL \url{https://ojs.aaai.org/index.php/AAAI/article/view/11757}.

\bibitem[Hiraoka et~al.(2022)Hiraoka, Imagawa, Hashimoto, Onishi, and
  Tsuruoka]{droq}
Hiraoka, T., Imagawa, T., Hashimoto, T., Onishi, T., and Tsuruoka, Y.
\newblock Dropout q-functions for doubly efficient reinforcement learning.
\newblock In \emph{International Conference on Learning Representations}, 2022.
\newblock URL \url{https://openreview.net/forum?id=xCVJMsPv3RT}.

\bibitem[Kalashnikov et~al.(2018)Kalashnikov, Irpan, Pastor, Ibarz, Herzog,
  Jang, Quillen, Holly, Kalakrishnan, Vanhoucke, and Levine]{qt-opt}
Kalashnikov, D., Irpan, A., Pastor, P., Ibarz, J., Herzog, A., Jang, E.,
  Quillen, D., Holly, E., Kalakrishnan, M., Vanhoucke, V., and Levine, S.
\newblock Scalable deep reinforcement learning for vision-based robotic
  manipulation.
\newblock In Billard, A., Dragan, A., Peters, J., and Morimoto, J. (eds.),
  \emph{Proceedings of The 2nd Conference on Robot Learning}, volume~87 of
  \emph{Proceedings of Machine Learning Research}, pp.\  651--673. PMLR, 29--31
  Oct 2018.
\newblock URL \url{https://proceedings.mlr.press/v87/kalashnikov18a.html}.

\bibitem[Kostrikov et~al.(2021)Kostrikov, Yarats, and Fergus]{drq}
Kostrikov, I., Yarats, D., and Fergus, R.
\newblock Image augmentation is all you need: Regularizing deep reinforcement
  learning from pixels.
\newblock In \emph{International Conference on Learning Representations}, 2021.
\newblock URL \url{https://openreview.net/forum?id=GY6-6sTvGaf}.

\bibitem[Kostrikov et~al.(2022)Kostrikov, Nair, and Levine]{iql}
Kostrikov, I., Nair, A., and Levine, S.
\newblock Offline reinforcement learning with implicit q-learning.
\newblock In \emph{International Conference on Learning Representations}, 2022.
\newblock URL \url{https://openreview.net/forum?id=68n2s9ZJWF8}.

\bibitem[Lee et~al.(2021)Lee, Seo, Lee, Abbeel, and Shin]{off2on}
Lee, S., Seo, Y., Lee, K., Abbeel, P., and Shin, J.
\newblock Offline-to-online reinforcement learning via balanced replay and
  pessimistic q-ensemble.
\newblock In \emph{5th Annual Conference on Robot Learning}, 2021.
\newblock URL \url{https://openreview.net/forum?id=AlJXhEI6J5W}.

\bibitem[Levine \& Koltun(2013)Levine and Koltun]{gps}
Levine, S. and Koltun, V.
\newblock Guided policy search.
\newblock In Dasgupta, S. and McAllester, D. (eds.), \emph{Proceedings of the
  30th International Conference on Machine Learning}, volume~28 of
  \emph{Proceedings of Machine Learning Research}, pp.\  1--9, Atlanta,
  Georgia, USA, 17--19 Jun 2013. PMLR.
\newblock URL \url{https://proceedings.mlr.press/v28/levine13.html}.

\bibitem[Levine et~al.(2020)Levine, Kumar, Tucker, and Fu]{offlineRL}
Levine, S., Kumar, A., Tucker, G., and Fu, J.
\newblock Offline reinforcement learning: Tutorial, review, and perspectives on
  open problems, 2020.
\newblock URL \url{https://arxiv.org/abs/2005.01643}.

\bibitem[Li et~al.(2022)Li, Kumar, Kostrikov, and Levine]{li2022efficient}
Li, Q., Kumar, A., Kostrikov, I., and Levine, S.
\newblock Efficient deep reinforcement learning requires regulating statistical
  overfitting.
\newblock In \emph{Deep Reinforcement Learning Workshop NeurIPS 2022}, 2022.
\newblock URL \url{https://openreview.net/forum?id=Jwfa-oyQduy}.

\bibitem[Lu et~al.(2022)Lu, Ball, Rudner, Parker-Holder, Osborne, and
  Teh]{lu2022challenges}
Lu, C., Ball, P.~J., Rudner, T. G.~J., Parker-Holder, J., Osborne, M.~A., and
  Teh, Y.~W.
\newblock Challenges and opportunities in offline reinforcement learning from
  visual observations.
\newblock In \emph{Workshop on Learning from Diverse, Offline Data}, 2022.
\newblock URL \url{https://openreview.net/forum?id=bPOBIKaqLba}.

\bibitem[Lu et~al.(2021)Lu, Hausman, Chebotar, Yan, Jang, Herzog, Xiao, Irpan,
  Khansari, Kalashnikov, and Levine]{awopt2021corl}
Lu, Y., Hausman, K., Chebotar, Y., Yan, M., Jang, E., Herzog, A., Xiao, T.,
  Irpan, A., Khansari, M., Kalashnikov, D., and Levine, S.
\newblock Aw-opt: Learning robotic skills with imitation andreinforcement at
  scale.
\newblock In \emph{5th Annual Conference on Robot Learning}, 2021.

\bibitem[Mirhoseini et~al.(2021)Mirhoseini, Goldie, Yazgan, Jiang, Songhori,
  Wang, Lee, Johnson, Pathak, Nazi, Pak, Tong, Srinivasa, Hang, Tuncer, Le,
  Laudon, Ho, Carpenter, and Dean]{Mirhoseini2021AGP}
Mirhoseini, A., Goldie, A., Yazgan, M., Jiang, J.~W., Songhori, E.~M., Wang,
  S., Lee, Y.-J., Johnson, E., Pathak, O., Nazi, A., Pak, J., Tong, A.,
  Srinivasa, K., Hang, W., Tuncer, E., Le, Q.~V., Laudon, J., Ho, R.,
  Carpenter, R., and Dean, J.
\newblock A graph placement methodology for fast chip design.
\newblock \emph{Nature}, 594 7862:\penalty0 207--212, 2021.

\bibitem[Mnih et~al.(2015)Mnih, Kavukcuoglu, Silver, Rusu, Veness, Bellemare,
  Graves, Riedmiller, Fidjeland, Ostrovski, Petersen, Beattie, Sadik,
  Antonoglou, King, Kumaran, Wierstra, Legg, and Hassabis]{mnih2015humanlevel}
Mnih, V., Kavukcuoglu, K., Silver, D., Rusu, A.~A., Veness, J., Bellemare,
  M.~G., Graves, A., Riedmiller, M., Fidjeland, A.~K., Ostrovski, G., Petersen,
  S., Beattie, C., Sadik, A., Antonoglou, I., King, H., Kumaran, D., Wierstra,
  D., Legg, S., and Hassabis, D.
\newblock Human-level control through deep reinforcement learning.
\newblock \emph{Nature}, 518\penalty0 (7540):\penalty0 529--533, February 2015.
\newblock ISSN 00280836.
\newblock URL \url{http://dx.doi.org/10.1038/nature14236}.

\bibitem[Moskovitz et~al.(2021)Moskovitz, Parker-Holder, Pacchiano, Arbel, and
  Jordan]{top}
Moskovitz, T., Parker-Holder, J., Pacchiano, A., Arbel, M., and Jordan, M.
\newblock Tactical optimism and pessimism for deep reinforcement learning.
\newblock In Ranzato, M., Beygelzimer, A., Dauphin, Y., Liang, P., and Vaughan,
  J.~W. (eds.), \emph{Advances in Neural Information Processing Systems},
  volume~34, pp.\  12849--12863. Curran Associates, Inc., 2021.
\newblock URL
  \url{https://proceedings.neurips.cc/paper/2021/file/6abcc8f24321d1eb8c95855eab78ee95-Paper.pdf}.

\bibitem[Nair et~al.(2018{\natexlab{a}})Nair, McGrew, Andrychowicz, Zaremba,
  and Abbeel]{ddpgBC}
Nair, A., McGrew, B., Andrychowicz, M., Zaremba, W., and Abbeel, P.
\newblock Overcoming exploration in reinforcement learning with demonstrations.
\newblock In \emph{2018 IEEE International Conference on Robotics and
  Automation (ICRA)}, pp.\  6292–6299. IEEE Press, 2018{\natexlab{a}}.
\newblock \doi{10.1109/ICRA.2018.8463162}.
\newblock URL \url{https://doi.org/10.1109/ICRA.2018.8463162}.

\bibitem[Nair et~al.(2018{\natexlab{b}})Nair, McGrew, Andrychowicz, Zaremba,
  and Abbeel]{nair2018overcoming}
Nair, A., McGrew, B., Andrychowicz, M., Zaremba, W., and Abbeel, P.
\newblock Overcoming exploration in reinforcement learning with demonstrations.
\newblock In \emph{2018 IEEE international conference on robotics and
  automation (ICRA)}, pp.\  6292--6299. IEEE, 2018{\natexlab{b}}.

\bibitem[Nair et~al.(2020)Nair, Gupta, Dalal, and Levine]{awac}
Nair, A., Gupta, A., Dalal, M., and Levine, S.
\newblock {AWAC}: Accelerating online reinforcement learning with offline
  datasets.
\newblock \emph{arXiv}, June 2020.

\bibitem[Ouyang et~al.(2022)Ouyang, Wu, Jiang, Almeida, Wainwright, Mishkin,
  Zhang, Agarwal, Slama, Gray, Schulman, Hilton, Kelton, Miller, Simens,
  Askell, Welinder, Christiano, Leike, and Lowe]{ouyang2022training}
Ouyang, L., Wu, J., Jiang, X., Almeida, D., Wainwright, C., Mishkin, P., Zhang,
  C., Agarwal, S., Slama, K., Gray, A., Schulman, J., Hilton, J., Kelton, F.,
  Miller, L., Simens, M., Askell, A., Welinder, P., Christiano, P., Leike, J.,
  and Lowe, R.
\newblock Training language models to follow instructions with human feedback.
\newblock In Oh, A.~H., Agarwal, A., Belgrave, D., and Cho, K. (eds.),
  \emph{Advances in Neural Information Processing Systems}, 2022.
\newblock URL \url{https://openreview.net/forum?id=TG8KACxEON}.

\bibitem[Rajeswaran et~al.(2018)Rajeswaran, Kumar, Gupta, Vezzani, Schulman,
  Todorov, and Levine]{dapg}
Rajeswaran, A., Kumar, V., Gupta, A., Vezzani, G., Schulman, J., Todorov, E.,
  and Levine, S.
\newblock {Learning Complex Dexterous Manipulation with Deep Reinforcement
  Learning and Demonstrations}.
\newblock In \emph{Proceedings of Robotics: Science and Systems (RSS)}, 2018.

\bibitem[Rezaeifar et~al.(2022)Rezaeifar, Dadashi, Vieillard, Hussenot, Bachem,
  Pietquin, and Geist]{offlineRLAntiExplore}
Rezaeifar, S., Dadashi, R., Vieillard, N., Hussenot, L., Bachem, O., Pietquin,
  O., and Geist, M.
\newblock Offline reinforcement learning as anti-exploration.
\newblock \emph{Proceedings of the AAAI Conference on Artificial Intelligence},
  36\penalty0 (7):\penalty0 8106--8114, Jun. 2022.
\newblock \doi{10.1609/aaai.v36i7.20783}.
\newblock URL \url{https://ojs.aaai.org/index.php/AAAI/article/view/20783}.

\bibitem[Ross \& Bagnell(2012)Ross and Bagnell]{rossbagnellagnostic}
Ross, S. and Bagnell, J.~A.
\newblock Agnostic system identification for model-based reinforcement
  learning.
\newblock In \emph{Proceedings of the 29th International Coference on
  International Conference on Machine Learning}, ICML'12, pp.\  1905–1912,
  Madison, WI, USA, 2012. Omnipress.
\newblock ISBN 9781450312851.

\bibitem[Rudner et~al.(2021)Rudner, Lu, Osborne, Gal, and Teh]{rudner2021on}
Rudner, T. G.~J., Lu, C., Osborne, M., Gal, Y., and Teh, Y.~W.
\newblock On pathologies in {KL}-regularized reinforcement learning from expert
  demonstrations.
\newblock In Beygelzimer, A., Dauphin, Y., Liang, P., and Vaughan, J.~W.
  (eds.), \emph{Advances in Neural Information Processing Systems}, 2021.
\newblock URL \url{https://openreview.net/forum?id=sS8rRmgAatA}.

\bibitem[Schaal(1996)]{LearningFromDemos}
Schaal, S.
\newblock Learning from demonstration.
\newblock In Mozer, M., Jordan, M., and Petsche, T. (eds.), \emph{Advances in
  Neural Information Processing Systems}, volume~9. MIT Press, 1996.
\newblock URL
  \url{https://proceedings.neurips.cc/paper/1996/file/68d13cf26c4b4f4f932e3eff990093ba-Paper.pdf}.

\bibitem[Silver et~al.(2016)Silver, Huang, Maddison, Guez, Sifre, van~den
  Driessche, Schrittwieser, Antonoglou, Panneershelvam, Lanctot, Dieleman,
  Grewe, Nham, Kalchbrenner, Sutskever, Lillicrap, Leach, Kavukcuoglu, Graepel,
  and Hassabis]{alphago}
Silver, D., Huang, A., Maddison, C.~J., Guez, A., Sifre, L., van~den Driessche,
  G., Schrittwieser, J., Antonoglou, I., Panneershelvam, V., Lanctot, M.,
  Dieleman, S., Grewe, D., Nham, J., Kalchbrenner, N., Sutskever, I.,
  Lillicrap, T., Leach, M., Kavukcuoglu, K., Graepel, T., and Hassabis, D.
\newblock Mastering the game of {Go} with deep neural networks and tree search.
\newblock \emph{Nature}, 529\penalty0 (7587):\penalty0 484--489, January 2016.
\newblock \doi{10.1038/nature16961}.

\bibitem[Song et~al.(2023)Song, Zhou, Sekhari, Bagnell, Krishnamurthy, and
  Sun]{song2023hybrid}
Song, Y., Zhou, Y., Sekhari, A., Bagnell, D., Krishnamurthy, A., and Sun, W.
\newblock Hybrid {RL}: Using both offline and online data can make {RL}
  efficient.
\newblock In \emph{International Conference on Learning Representations}, 2023.
\newblock URL \url{https://openreview.net/forum?id=yyBis80iUuU}.

\bibitem[Srivastava et~al.(2014)Srivastava, Hinton, Krizhevsky, Sutskever, and
  Salakhutdinov]{dropout}
Srivastava, N., Hinton, G., Krizhevsky, A., Sutskever, I., and Salakhutdinov,
  R.
\newblock Dropout: A simple way to prevent neural networks from overfitting.
\newblock \emph{Journal of Machine Learning Research}, 15\penalty0
  (56):\penalty0 1929--1958, 2014.
\newblock URL \url{http://jmlr.org/papers/v15/srivastava14a.html}.

\bibitem[Thrun \& Schwartz(1993)Thrun and Schwartz]{thrunOverestimate}
Thrun, S. and Schwartz, A.
\newblock Issues in using function approximation for reinforcement learning.
\newblock In \emph{Proceedings of 4th Connectionist Models Summer School}.
  Erlbaum Associates, June 1993.

\bibitem[van Hasselt et~al.(2016)van Hasselt, Guez, and
  Silver]{Hasselt2015DeepRL}
van Hasselt, H., Guez, A., and Silver, D.
\newblock Deep reinforcement learning with double q-learning.
\newblock \emph{Proceedings of the AAAI Conference on Artificial Intelligence},
  30\penalty0 (1), Mar. 2016.
\newblock \doi{10.1609/aaai.v30i1.10295}.
\newblock URL \url{https://ojs.aaai.org/index.php/AAAI/article/view/10295}.

\bibitem[Ve{\v c}er{\'\i}k et~al.(2017)Ve{\v c}er{\'\i}k, Hester, Scholz, Wang,
  Pietquin, Piot, Heess, Roth{\"o}rl, Lampe, and Riedmiller]{ddpgfd}
Ve{\v c}er{\'\i}k, M., Hester, T., Scholz, J., Wang, F., Pietquin, O., Piot,
  B., Heess, N., Roth{\"o}rl, T., Lampe, T., and Riedmiller, M.
\newblock Leveraging demonstrations for deep reinforcement learning on robotics
  problems with sparse rewards.
\newblock \emph{arXiv}, July 2017.

\bibitem[Wagenmaker \& Pacchiano(2022)Wagenmaker and
  Pacchiano]{wagenmaker2022leveraging}
Wagenmaker, A. and Pacchiano, A.
\newblock Leveraging offline data in online reinforcement learning.
\newblock \emph{arXiv preprint arXiv:2211.04974}, 2022.

\bibitem[Yarats et~al.(2022)Yarats, Fergus, Lazaric, and
  Pinto]{yarats2022mastering}
Yarats, D., Fergus, R., Lazaric, A., and Pinto, L.
\newblock Mastering visual continuous control: Improved data-augmented
  reinforcement learning.
\newblock In \emph{International Conference on Learning Representations}, 2022.
\newblock URL \url{https://openreview.net/forum?id=_SJ-_yyes8}.

\bibitem[Zhang et~al.(2023)Zhang, Xu, and Yu]{zhang2023policy}
Zhang, H., Xu, W., and Yu, H.
\newblock Policy expansion for bridging offline-to-online reinforcement
  learning.
\newblock In \emph{The Eleventh International Conference on Learning
  Representations}, 2023.
\newblock URL \url{https://openreview.net/forum?id=-Y34L45JR6z}.

\end{thebibliography}
\bibliographystyle{icml2023}

\newpage
\appendix
\onecolumn

\section{Detailed Experiments}
\label{sec:detailedexperiments}

\subsection{Sparse Adroit}

\paragraph{Full Results} Here we present the full results for Sparse Adroit, including LayerNorm and environment-specific design choice ablations.

\begin{figure}[H]
    \centering
    \includegraphics[width=0.7\textwidth]{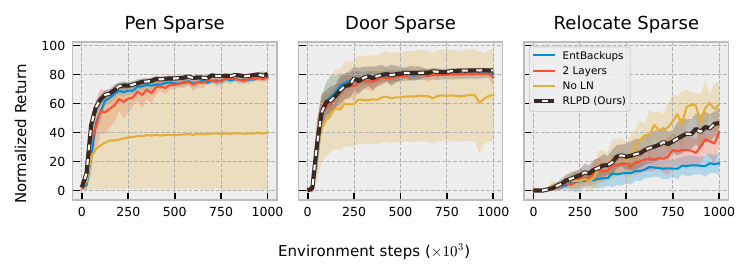} \\
    \vspace{-5mm}
    \caption{Full Adroit Results.}
    \label{fig:adroit-full}
\end{figure}

We see that LayerNorm greatly reduces variance on the Pen and Door environments, however slightly harms mean performance on Relocate. Taken together however, LayerNorm is still a vital ingredient for reliable performance on this domain.

\subsection{AntMaze}

\paragraph{Full Results} Here we present the full results for AntMaze, including LayerNorm and environment-specific design choice ablations.

\begin{figure}[H]
    \centering
    \includegraphics[width=0.7\textwidth]{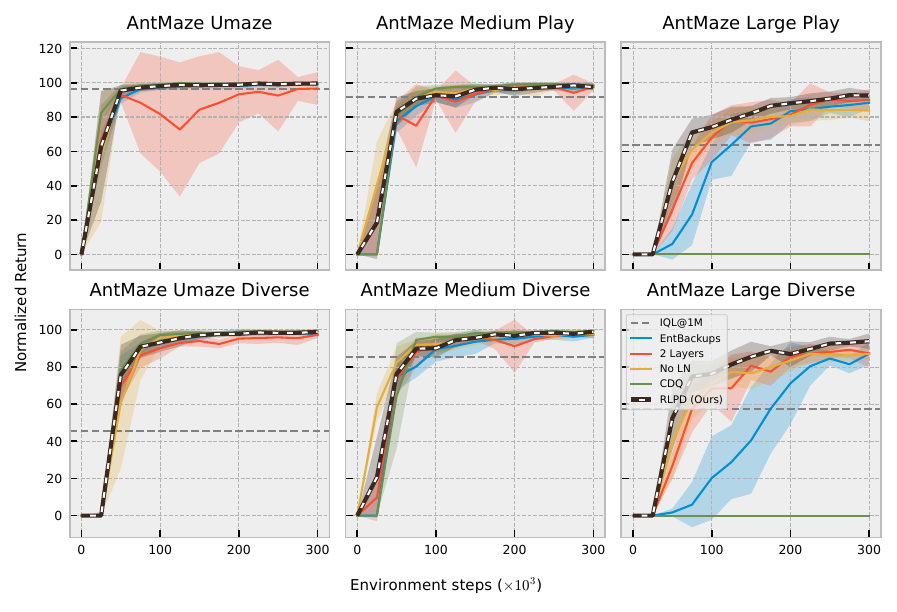} \\
    \vspace{-5mm}
    \caption{Full AntMaze results}
    \label{fig:antmaze-full}
\end{figure}

We see that our recommended design choices are vital for strong performance on the harder Large tasks, and furthermore see that using deeper 3-layer networks seems to help stability across all tasks. As we see, the best design choices on the Large environments are also optimal for the Umaze and Medium environments.
\newpage
\paragraph{Gradient Steps per Environment Step Ablations}

\begin{figure}[H]
    \centering
    \includegraphics[width=0.7\textwidth]{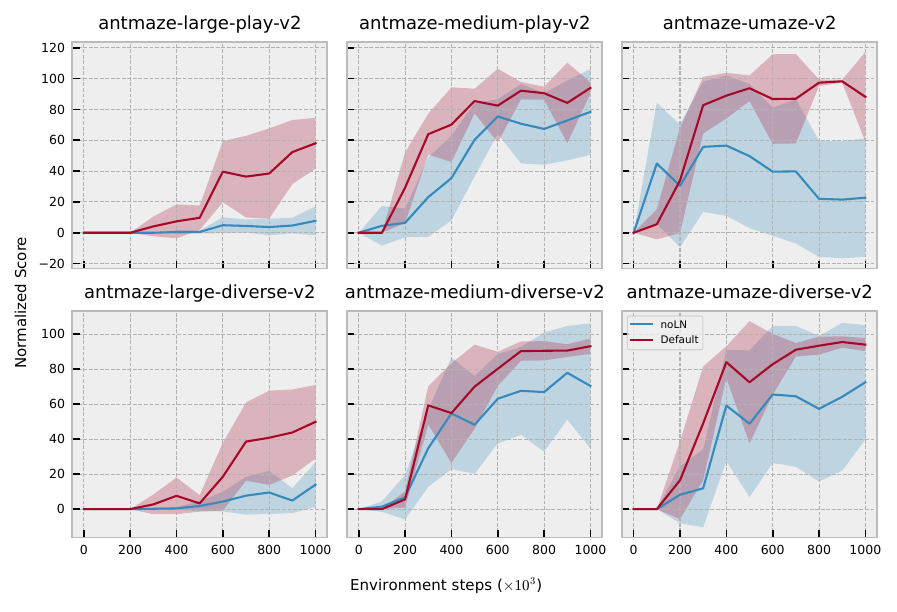} \\
    \vspace{-5mm}
    \caption{SAC with and without Layer Normalization.}
    \label{fig:antmaze-sac}
\end{figure}

We see LayerNorm is incredibly effective in a setting whereby we perform a single gradient update per time-step in \cref{fig:antmaze-sac}. This may be required for learning on real robots, where computational efficiency is important.

\begin{figure}[H]
    \centering
    \includegraphics[width=0.7\textwidth]{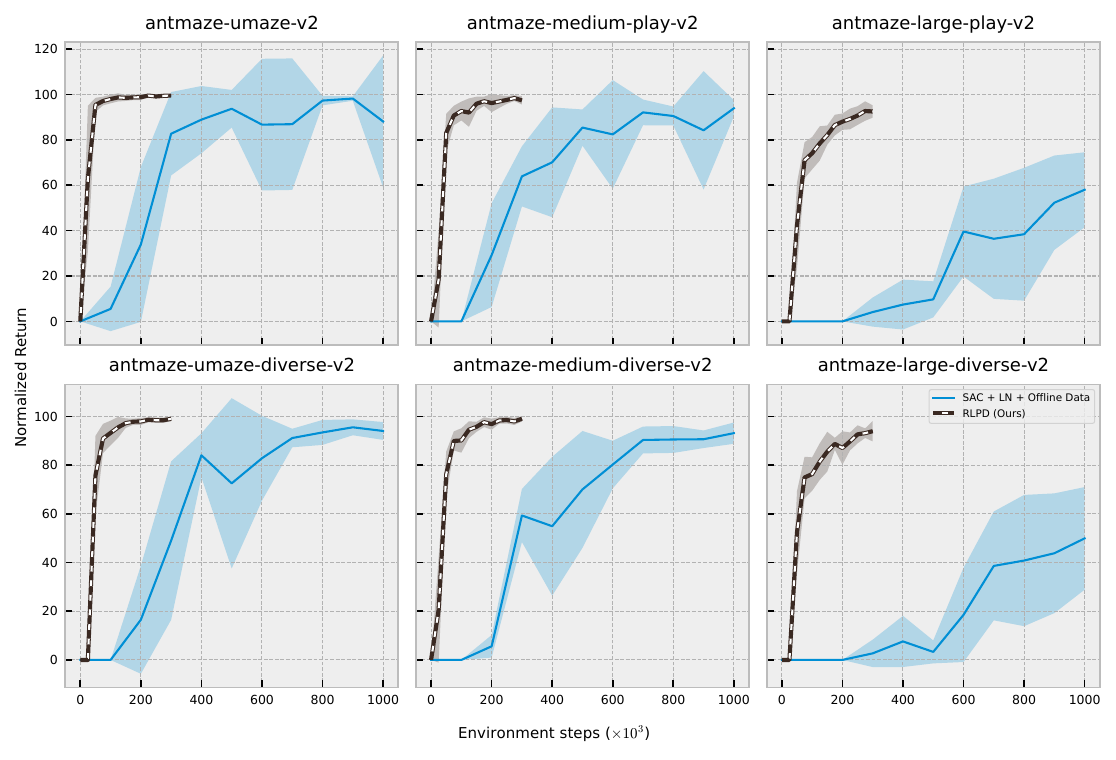} \\
    \vspace{-5mm}
    \caption{\ouralgo v.s. SAC + LN + Offline Data.}
    \label{fig:antmaze-utd}
\end{figure}

In \cref{fig:antmaze-utd}, we see the impact of increasing the number of gradient steps per time-step. As we see, both sample efficiency stability improves greatly.

\newpage
\paragraph{Full IQL comparison}

\begin{figure}[H]
    \centering
    \includegraphics[width=0.7\textwidth]{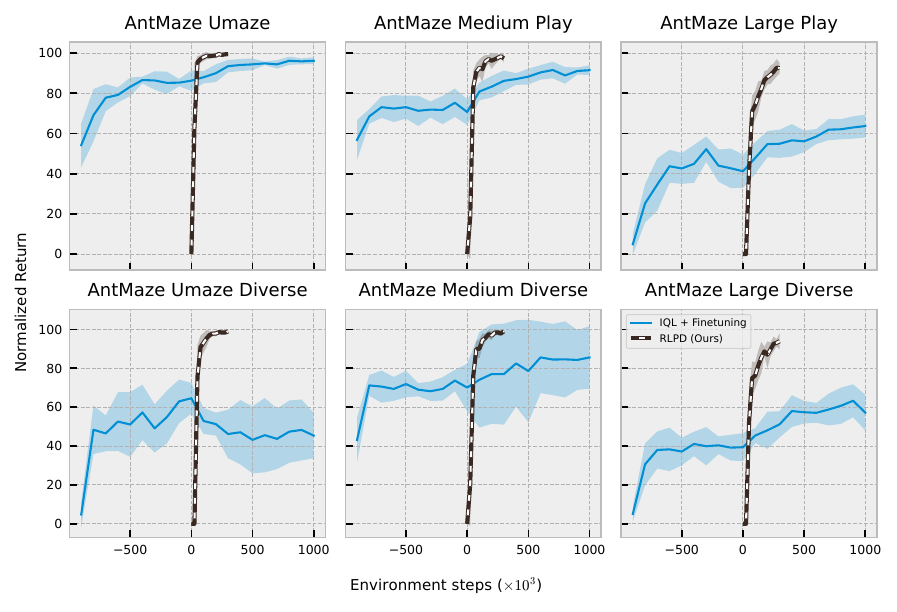} \\
    \vspace{-5mm}
    \caption{\ouralgo v.s. IQL + Finetuning. We show offline pre-training steps for IQL using negative indices.}
    \label{fig:antmaze-iql-compare}
\end{figure}

Here we compare to IQL + Finetuning. We show pre-training gradient steps on IQL as negative indices in the x-axis. As we see in \cref{fig:antmaze-iql-compare}, despite IQL learning a strong initialization, it struggles to improve beyond this. In comparison, \ouralgo is able to quickly match, then greatly exceed IQL.

\subsection{D4RL Locomotion}

\paragraph{Full Results} Here we present the full results for D4RL Locomotion, including LayerNorm and baselines. We emphasize that these tasks are not necessarily a good use case for online learning with offline datasets, as the tasks can be solved relatively quickly with purely online approaches, and there is no inherent difficulty regarding exploration.

\begin{figure}[H]
    \centering
    \includegraphics[width=0.7\textwidth]{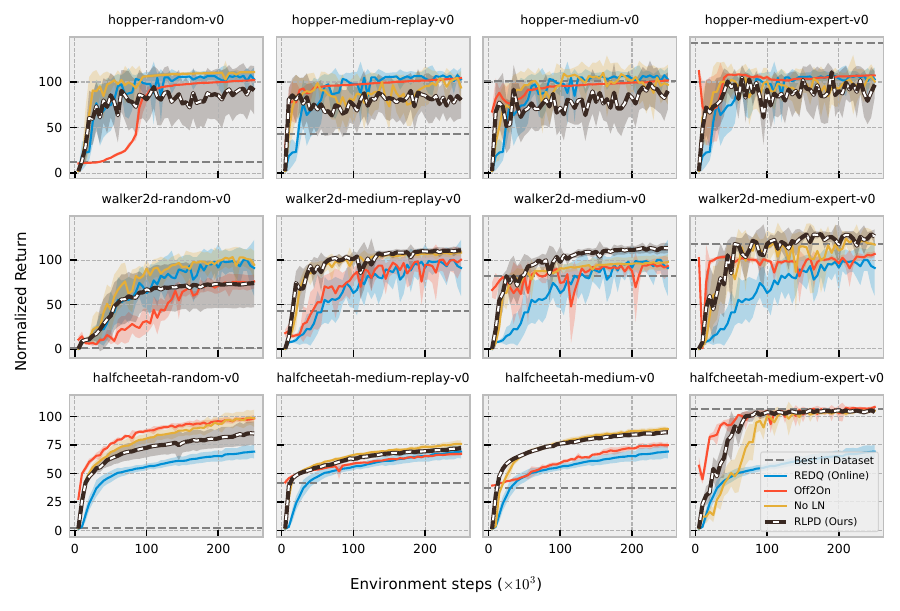}
    \vspace{-5mm}
    \caption{D4RL Ablations.}
    \label{fig:locoablations}
    \vspace{-4mm}
\end{figure}

The impact of LayerNorm is not so clear cut in \cref{fig:locoablations}; this is to be expected as online approaches already achieve strong results in this domain. Notably, we see strong performance compared to baselines in the medium-expert domains.

\begin{figure}[H]
    \vspace{10mm}
    \centering
    \includegraphics[width=0.7\textwidth]{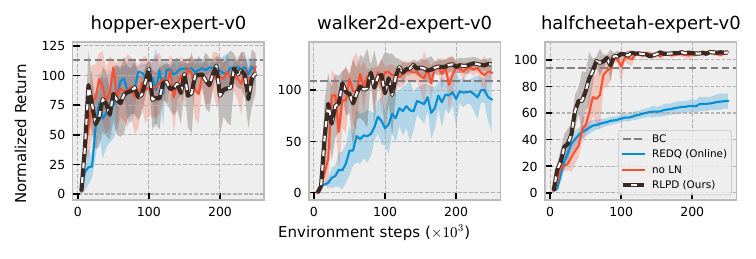}
    \vspace{-5mm}
    \caption{D4RL Ablation on Expert Data.}
    \label{fig:locoexpert}
    \vspace{-5mm}
\end{figure}

We also test on the narrow Expert dataset, not tested by \citep{off2on}. In \cref{fig:locoexpert} we see LayerNorm can marginally help both sample efficiency, and asymptotic performance.

\subsection{V-D4RL Locomotion}

\paragraph{Full Results} Here we present the full results for V-D4RL Locomotion, including LayerNorm and environment-specific design choice ablations.

\begin{figure}[H]
    \centering
    \includegraphics[width=0.7\textwidth]{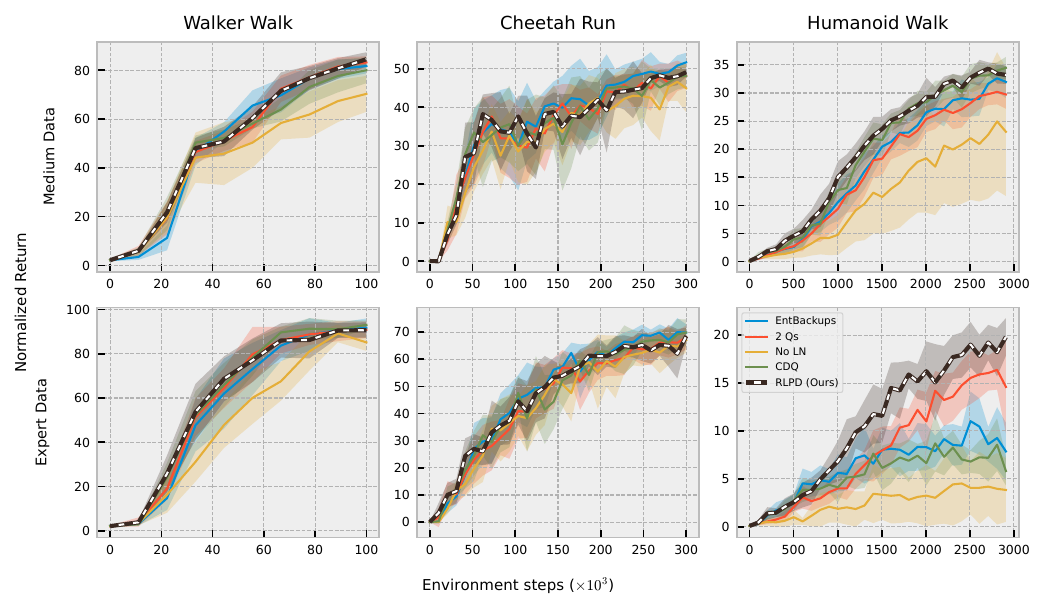}
    \vspace{-5mm}
    \caption{VD4RL Ablations.}
    \label{fig:vd4rlablations}
\end{figure}

As we see, LayerNorm helps significantly in the Walker and Humanoid environments. We also see the positive impact of our recommended design choices in the complex Humanoid domain.

\newpage
\section{Experimental Details}

\subsection{Further Environment Details}
\label{sec:envdetails}

\begin{figure}[H]
    \centering
    \begin{subfigure}[b]{0.9\textwidth}
    \centering
    \includegraphics[width=0.7\textwidth]{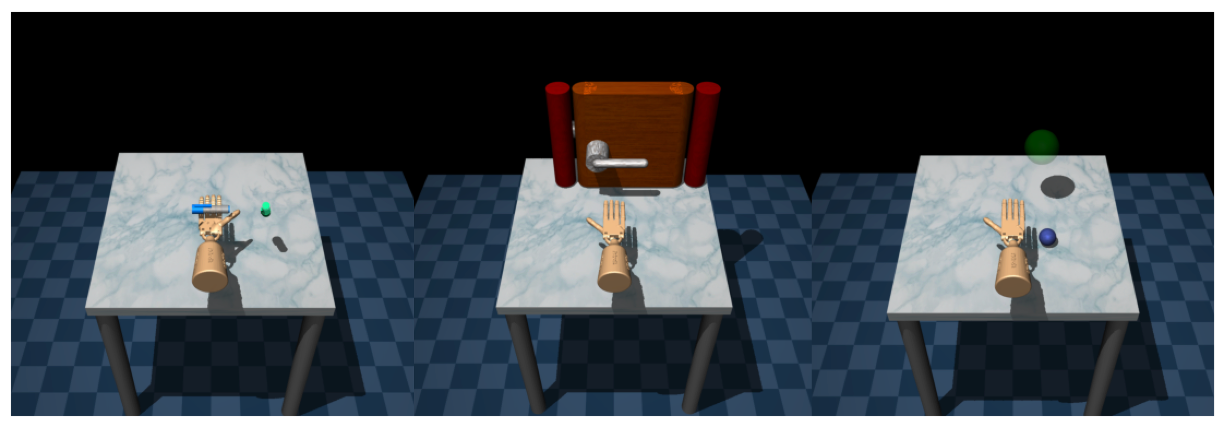}
    \vspace{-3mm}
    \caption{The Sparse Adroit Domain. Pen, Door and Relocate tasks respectively.}
    \end{subfigure}
    \begin{subfigure}[b]{0.9\textwidth}
    \centering
    \includegraphics[width=0.7\textwidth]{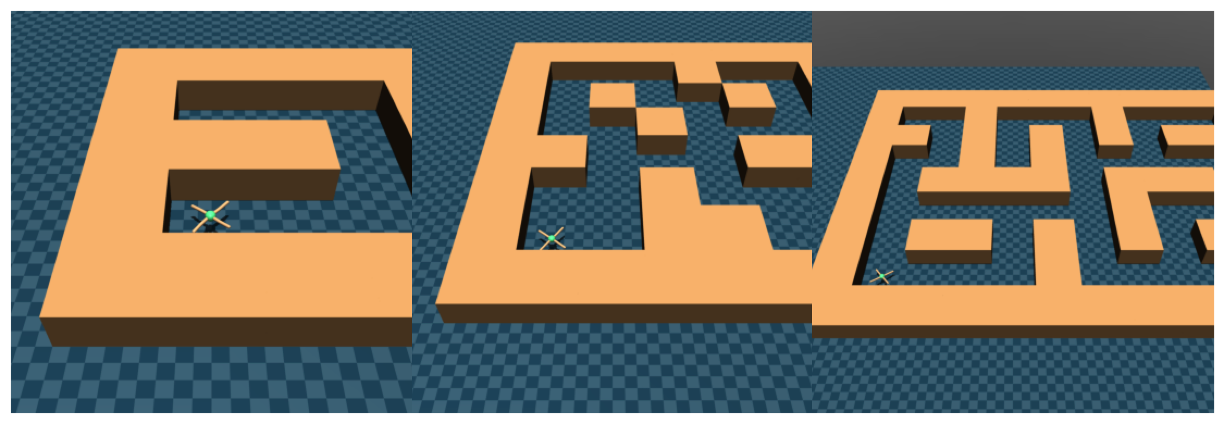}
    \vspace{-3mm}
    \caption{The AntMaze Domain. Umaze, Medium and Large tasks respectively.}
    \end{subfigure}
    \begin{subfigure}[b]{0.9\textwidth}
    \centering
    \includegraphics[width=0.7\textwidth]{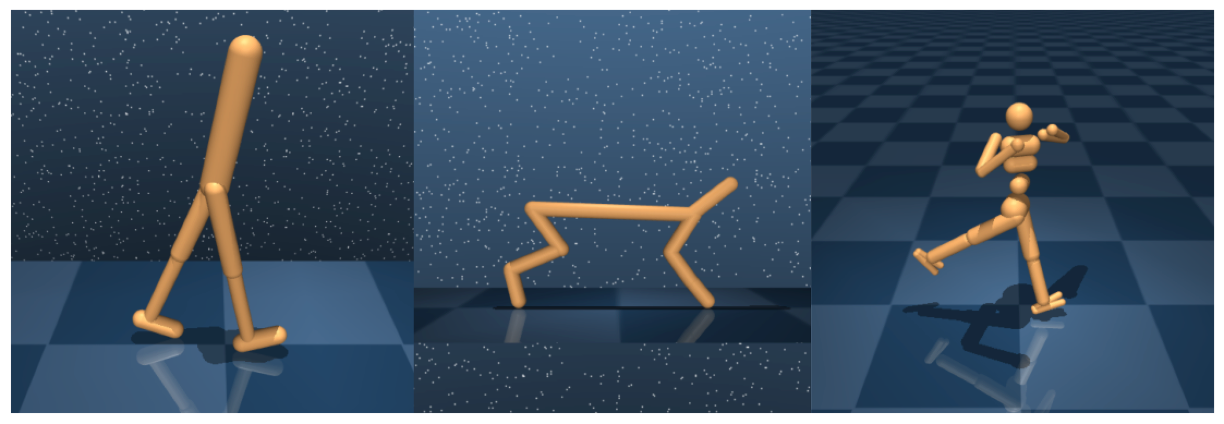}
    \vspace{-3mm}
    \caption{The V-D4RL Domain. Walker Walk, Cheetah Run and Humanoid Walk respectively.}
    \end{subfigure}
    \vspace{-2mm}
    \caption{Visualizations of the environments we consider.}
    \label{fig:envs}
\end{figure}

We provide further details about the key domains we evaluate on. In \cref{fig:envs} we provide visualizations of the environments.

\paragraph{Sparse Adroit}
In these tasks, reward is a binary variable that indicates whether the task has been completed successfully or not. In prior work \citep{awac, rudner2021on}, it is common to see success rate used as the metric, which simply determines if the task has been completed in any time step. However \citet{iql} use a more challenging metric that involves speed of completion. Concretely, return is calculated as the percentage of the total timesteps in which the task is considered solved (note that there are no early terminations). For example, in the `Pen' task, where the horizon is 100 timesteps, if a policy achieves a Normalized Score of 80, that means in 80 timesteps, the task is considered solved. This effectively means that the policy was able to solve the task in 20 timesteps. At each evaluation, we perform 100 trials.

\paragraph{D4RL: AntMaze} In these tasks, reward is a binary variable that indicates whether the agent has reached the goal. Upon reaching the goal, the episode terminates. The normalized return is therefore the proportion of evaluation trials that were successful. We follow prior work, and perform 100 trials, and measure Normalized Return as the percentage of successful trials.

\newpage
\subsection{Hyperparameters}

Here we list the hyperparameters for \ouralgo in \cref{tab:rlpd_hyperparameters}, and the environment-specific hyperparameters in \cref{tab:env_hyperparameters}.

\begin{table}[ht!]
\centering
\caption{\ouralgo hyperparameters.}
\label{tab:rlpd_hyperparameters}
\begin{tabular}{l|c}
\toprule
\textbf{Parameter}                    & \textbf{Value}   \\ \hline
Online batch size                            & 128                   \\
Offline batch size                            & 128                   \\
Discount ($\gamma$)                             & 0.99              \\
Optimizer                             & Adam              \\
Learning rate                             & $3\times 10^{-4}$              \\
Ensemble size ($E$)                       & 10 \\
Critic EMA weight ($\rho$)          & 0.005 \\
Gradient Steps (State Based) ($G$ or UTD)   &  20 \\
Network Width                        & 256 Units \\
Initial Entropy Temperature ($\alpha$) & 1.0 \\
Target Entropy & $-\dim(\mathcal{A}) / 2$ \\
\midrule
\multicolumn{2}{c}{\textbf{Pixel-Based Hyperparameters}} \\
\midrule
Action repeat                         & 2                   \\
Observation size                      & [64, 64]              \\
Image shift amount                    & 4                      \\
\bottomrule
\end{tabular}
\end{table}

\begin{table}[ht!]
\centering
\caption{Environment specific hyperparameters.}
\label{tab:env_hyperparameters}
\begin{tabular}{l|c|c|c}
\toprule
\textbf{Environment}                    & \textbf{CDQ} & \textbf{Entropy Backups} & \textbf{MLP Architecture}   \\ \hline
Locomotion & True & True & 2 Layer \\
AntMaze & False & False & 3 Layer \\
Adroit & True & False & 3 Layer \\
DMC (Pixels) & False & False & 2 Layer \\
\bottomrule
\end{tabular}
\end{table}


\end{document}